\documentclass{article}

\usepackage{arxiv}

\usepackage[utf8]{inputenc} 
\usepackage[T1]{fontenc}    
\usepackage{hyperref}       
\usepackage{url}            
\usepackage{booktabs}       
\usepackage{amsfonts}       
\usepackage{nicefrac}       
\usepackage{microtype}      
\usepackage{lipsum}		
\usepackage{graphicx}
\usepackage{natbib}
\usepackage{doi}
\usepackage{wrapfig}
\usepackage{float}
\usepackage{longtable}
\usepackage[table]{xcolor}

\title{The Thousand Brains Project: A New Paradigm for Sensorimotor Intelligence}

\author{ \href{https://orcid.org/0000-0001-9152-0666}{\includegraphics[scale=0.06]{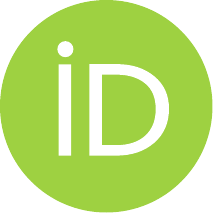}\hspace{1mm}Viviane ~Clay}\thanks{Joint first authors.} \\
	Numenta, Inc.\\
	Redwood City\\
	CA, United States \\
	\texttt{vclay@thousandbrains.org} \\
	\And
	\href{https://orcid.org/0000-0002-2540-7486}{\includegraphics[scale=0.06]{orcid.pdf}\hspace{1mm}Niels ~Leadholm*} \\
	Numenta, Inc.\\
	Redwood City\\
	CA, United States \\
	\texttt{nleadholm@thousandbrains.org} \\
	\And
	\href{}{\hspace{1mm}Jeff ~Hawkins} \\
	Numenta, Inc.\\
	Redwood City\\
	CA, United States \\
	\texttt{jhawkins@thousandbrains.org} \\
}

\renewcommand{\headeright}{}
\renewcommand{\undertitle}{}  
\renewcommand{\shorttitle}{The Thousand Brains Project}

\hypersetup{
pdftitle={Clay et al 2024},
pdfsubject={q-bio.NC, q-bio.QM},
pdfauthor={Viviane ~Clay, Niels ~Leadholm, Jeff ~Hawkins},
pdfkeywords={Sensorimotor, Neocortex, Embodied, General Intelligence, Reference Frames, Spatial Representations, Model-Based,
World Models, Canonical Microcircuit},
}

\begin{document}
\maketitle

\begin{abstract}
	Artificial intelligence has advanced rapidly in the last decade, driven primarily by progress in the scale of deep-learning systems. Despite these advances, the creation of intelligent systems that can operate effectively in diverse, real-world environments remains a significant challenge. In this white paper, we outline the Thousand Brains Project, an ongoing research effort to develop an alternative, complementary form of AI, derived from the operating principles of the neocortex. We present an early version of a thousand-brains system, a sensorimotor agent that is uniquely suited to quickly learn a wide range of tasks and eventually implement any capabilities the human neocortex has. Core to its design is the use of a repeating computational unit, the `learning module', modeled on the cortical columns found in mammalian brains. Each learning module operates as a semi-independent unit that can model entire objects, represents information through spatially structured reference frames, and both estimates and is able to effect movement in the world. Learning is a quick, associative process, similar to Hebbian learning in the brain, and leverages inductive biases around the spatial structure of the world to enable rapid and continual learning. Multiple learning modules can interact with one another both hierarchically and non-hierarchically via a \textit{cortical messaging protocol} (CMP), creating more abstract representations and supporting multimodal integration. We outline the key principles motivating the design of thousand-brains systems and provide details about the implementation of Monty, our first instantiation of such a system. Code can be found at \url{https://github.com/thousandbrainsproject/tbp.monty}, along with more detailed documentation at \url{https://thousandbrainsproject.readme.io/}.
\end{abstract}


\twocolumn
\section{Introduction}

We are developing a platform for building AI and robotics applications using the same principles as the human brain, a broad research initiative called the Thousand Brains Project. The principles this project builds on are fundamentally different from those used in deep learning, currently the most prevalent form of AI. Therefore, our platform represents an alternative form of AI, one that we believe will play an ever-increasing role in the future.

This paper outlines the motivation of the Thousand Brains Project, as well as the technical details of the underlying algorithm for sensorimotor intelligence. The aim is to enable developers to build AI applications that are more intelligent, more flexible, and more capable in the many applications that deep learning methods fail. Core to the design of thousand-brains systems are the principles laid out in the Thousand Brains Theory \citep{Hawkins2019ANeocortex}, a theory of intelligence derived from neuroscientific evidence of the anatomy and function of the neocortex. One core principle of the theory builds on the work of Vernon Mountcastle, who argued that the power of the mammalian brain lies in its re-use of cortical columns as the primary computational unit \citep{Mountcastle1997TheNeocortex, Edelman1978MindfulBrain}. In honor of Mountcastle's idea, we name the first practical implementation of a thousand brains system "Monty". The code for building and experimenting with Monty can be found at \url{https://github.com/thousandbrainsproject/tbp.monty}.

One key differentiator between thousand-brains systems and other AI technologies is that the former are built with embodied, sensorimotor learning at their core. Sensorimotor systems learn by sensing different parts of the world over time while interacting with it. For example, as you move your body, your limbs, and your eyes, the input to your brain changes. In thousand-brains systems, the learning derived from continuous interaction with an environment represents the foundational knowledge that supports all other functions. This contrasts with the growing approach that sensorimotor interactions are a sub-problem that can be solved by beginning with an architecture trained on a mixture of internet-scale language and multi-media data \citep{Driess2023, OpenAI2023gpt4, Black2024}. In addition to sensorimotor interaction being the core basis for learning, the centrality of sensorimotor learning manifests in the design choice that all levels of processing are sensorimotor. As will become clear, sensory and motor processing are not broken up and handled by distinct architectures, or limited to a single, global action output \citep{Reed2022, Driess2023, SIMAteam, Black2024}. Instead, sensation and motor outputs play a crucial role at every point in thousand-brains systems where information is processed.

A second differentiator is that our sensorimotor systems learn structured models, using \textit{reference frames}, explicit coordinate systems within which locations and rotations can be represented. Internal models derived from these reference frames keep track of where their sensors are relative to things in the world. Models are learned by assigning sensory observations to locations in reference frames. In this way, the models learned by sensorimotor systems are structured, similar to CAD models in a computer. This allows the system to quickly learn the structure of the world and how to manipulate objects to achieve a variety of goals, what is sometimes referred to as a \textit{world model}. As with sensorimotor learning, reference frames are used throughout all levels of information processing, including the representations of not only environments but also physical objects and abstract concepts - even the simplest representations are represented within a reference frame.

There are numerous advantages to sensorimotor learning and reference frames. At a high level, you can think about all the ways humans are different from today's AI. We learn quickly and continuously, constantly updating our knowledge of the world as we go about our day. We do not have to undergo a lengthy and expensive training phase to learn something new. We interact with the world and manipulate tools and objects in sophisticated ways that leverage our knowledge of how things are structured. For example, we can explore a new app on our phone and quickly figure out what it does and how it works based on other apps we know. We actively test hypotheses to fill in the gaps in our knowledge. We also learn from multiple modalities and these different sensory inputs work together seamlessly. For example, we may learn what a new tool looks like with a few glances and then immediately know how to grab and interact with the object via touch. Finally, we carry out complex, planned actions that leverage our knowledge of the world to enable intelligent behavior in new settings.

One of the most important discoveries about the brain is that most of what we think of as intelligence, from seeing, to touching, to hearing, to conceptual thinking, to language, is enabled by a common neural algorithm \citep{Mountcastle1997TheNeocortex}. All aspects of intelligence are created by the same sensorimotor mechanism. In the neocortex, this mechanism is implemented in each of the thousands of cortical columns. This means we can create many different types of intelligent systems using a set of common building blocks. The architecture we are creating is built on this premise. Thousand-brains systems will provide the core components and developers will then be able to assemble widely varying AI and robotics applications using them in different numbers and arrangements. We now elaborate on the high-level motivations of the Thousands Brains Project (TBP), before describing the technical details of Monty, the first instance of a thousand-brains system.

\section{The Thousand Brains Project}

\subsection{Long Term Goals} 

A central long-term goal is to build a universal platform and messaging protocol for intelligent sensorimotor systems. We call this protocol the "Cortical Messaging Protocol" (CMP). The CMP can be used as an interface between different \textit{modules}, and its universality is central to the ease of use of the SDK we are developing. For instance, one person may have modules optimized for flying a drone using birds-eye observations, while another may be working with different sensors and actuators regulating a smart home. Drone operation and smart-home control are quite different settings, but the modules used in these settings should be able to communicate through the same channels defined here. Furthermore, a setup for a larger home with multiple drones might require more modules to fully learn and control the system. The CMP is designed to enable rapid scaling of thousand-brains systems as required. Finally, third parties could develop sensor modules and learning modules (terms which we will shortly define) according to their specific requirements, but they would be compatible with all existing modules due to the shared messaging protocol.

A second goal of the TBP is to be a catalyst for a whole new way of thinking about machine intelligence. The principles of the TBP differ from many principles of popular AI methodologies today and are more in line with the principles of learning in the brain. Most concepts presented here derive from the Thousand Brains Theory (TBT) \citep{Hawkins2019ANeocortex} and experimental evidence about how the brain works. Modules in thousand-brains systems are inspired by cortical columns in the neocortex \citep{Edelman1978MindfulBrain, Mountcastle1997TheNeocortex}. The CMP between modules relies on object ID and pose information, as could be encoded in neural activity. The communication process is analogous to long-range connections in the neocortex.

In our implementation, we do not need to strictly adhere to all biological details, and it is important to note that should an engineering solution serve us better for implementing certain aspects, then it is acceptable to deviate from the neuroscience. For instance, we do not need to simulate spikes and can implement the general algorithm to be efficient on today's hardware. In general, the inner workings of the modules can can vary in implementation detail and do not have to rely on neuroscience as long as they adhere to the CMP. However, the core principles of the TBP are motivated by what we have learned from studying the neocortex. Furthermore, we expect that in many instances, the TBP will bring together prior work into a single framework, including sparsity, active dendrites, sequence memory, and grid cells \citep{Hawkins2016WhyNeocortex, Hawkins2017, Ahmad2019HowRepresentations, Hawkins2019ANeocortex, Lewis2019LocationsCells}.

Finally, it will be important to showcase the capabilities of our approach. We will work towards creating non-trivial demos where the implementation showcases capabilities that would be hard to demonstrate any other way. This may not be one specific task but could play to the strength of this system to tackle a wide variety of tasks. We will also work on making Monty an easy-to-use open-source SDK that other practitioners can apply and test on their applications. We want this to be a platform for all kinds of sensorimotor applications and not just a specific technology showcase.

\subsection{Core Principles}
We have a set of guiding principles that steer the Thousand Brains Project. Throughout the life of the project, there may be several different implementations, and within each implementation, there may be different versions of the core building blocks, but everything we work on should follow these core principles:
\begin{itemize}
  \setlength\itemsep{-0.1em}
  \item \textbf{Sensorimotor learning and inference:} We use actively generated temporal sequences of sensory and motor inputs instead of static inputs. The outputs of the system are motor commands.
  \item \textbf{Modular structure:} The same algorithm needs to work for all modalities. This general algorithm embodied in a learning module makes the system easily expandable and scalable.
  \item \textbf{Cortical Messaging Protocol:} The inputs and outputs of a learning module adhere to a defined protocol such that many different sensor modules (and modalities) and learning modules can work together seamlessly.
  \item \textbf{Voting:} A mechanism by which a collection of experts can use different information and models to come to a faster, more robust and stable conclusion. 
  \item \textbf{Reference frames:} The learned models should have inductive biases that make them naturally good at modeling a structured world that evolves over time. The learned models can be used for a variety of tasks such as manipulation, planning, imagining previously unseen states of the world, fast learning, generalization, and many more.
  \item \textbf{Rapid, continual learning where learning and inference are closely intertwined:} Supported by sensorimotor embodiment and reference frames, biologically plausible learning mechanisms enable rapid knowledge accumulation and updates to stored representations while remaining robust under the setting of continual learning. There is also no clear distinction between learning and inference. We are always learning, and always performing inference.
  \item \textbf{Model-free and model-based policies:} Low-level, model-free policies provide efficient means of interacting with the world,  but are crucially combined with model-based policies that support flexible action planning in novel situations.
\end{itemize}

In the initial implementation presented here, many components are deliberately \textit{not} biologically constrained, and/or simplified, so as to support visualizing, debugging, and understanding the system as a whole. For example, object models are currently based on explicit graphs in 3D Cartesian space. In the future, these elements may be substituted with more powerful, albeit more inscrutable neural components.

\subsection{Challenging Preconceptions}

Several of the ideas and ways of thinking introduced in this document may be counter-intuitive to those familiar with current AI methods, including deep learning. For example, ideas about intelligent systems, learning, models, hierarchical processing, or action policies that you already have in mind might not apply to the system that we are describing. We therefore ask the reader to try and dispense with as many preconceptions as possible and to understand the ideas presented here on their own terms. We are happy to discuss any questions or thoughts that may arise from reading this document. Please join our \href{https://thousandbrains.discourse.group/}{Discourse forum} or reach out to us at \href{mailto:info@thousandbrains.org}{info@thousandbrains.org}.

Below, we highlight some of the most important differences between the system we are building and other AI systems.

\begin{itemize}
  \item We are building a sensorimotor system. It learns by interacting with the world and sensing different parts of it over time. It \textbf{does not learn from a static dataset}. This is a fundamentally different way of learning than most leading AI systems today and addresses a (partially overlapping) different set of problems.
  \item We will introduce learning modules as the basic, repeatable modeling unit, comparable to a cortical column. An important detail to point out here is that none of these modeling units receives the full sensory input. For example, in vision, \textbf{there is no `full image' anywhere}. Each sensor senses a small patch in the world. This is in contrast to many AI systems today, where all sensory input is fed into a single model.
  \item Despite the previous point, \textbf{each modeling system can learn complete models of objects} and recognize them on its own. \textbf{A single modeling unit should be able to perform all basic tasks of object recognition and manipulation}. Using more modeling units makes the system faster and more efficient and supports compositional and abstract representations, but a single learning module is itself a powerful system. In the single module scenario, inference always requires movement to collect a series of observations, in the same way that recognizing a coffee cup with one finger requires moving across its surface.
  \item All models are structured by reference frames. An object is not just a bag of features. It is a collection of features at locations. \textbf{The relative locations of features to each other are more important than the features themselves}. These principles are used for modeling all discrete concepts in the world, from the simplest of physical objects to abstract concepts in society or mathematics.
  \item \textbf{Action policies are, first and foremost, model-based}. Learned models of objects in the world are used to determine appropriate actions in novel situations. Any given learning module can use its internal models to propose goal-states that are either decomposed into simpler goal-states in other learning modules, or are acted upon directly by motor systems. In this way, complex policies can be hierarchically decomposed, while still leveraging learned models. Over time and with practice, model-based policies become more efficient, while model-free policies can learn to do certain tasks rapidly and with finesse, but model-free policies are not the initial basis of actions in unfamiliar settings.
  \item \textbf{Motor output can be generated at any level of the system.} In contrast to many current approaches for sensorimotor interaction, we do not have a separate hierarchy of sensory processing followed by the generation of motor commands. Instead, each learning module, even at the lowest sensory level, produces action outputs. This is analogous to the projections to subcortical motor regions found in every area of the neocortex, even regions classically thought of as sensory regions.
\end{itemize}

\subsection{Capabilities of the System}
The thousand-brains architecture is designed to be a general-purpose AI system. It is not designed to solve a specific task or set of tasks. Instead, it is designed to be a platform that can be used to build a wide variety of AI applications. Like an operating system or a programming language does not define what the user applies it to, the Thousand Brains Project will provide the tools necessary to solve many of today’s current problems as well as completely new and unanticipated ones without being specific to any one of them.

Even though we cannot predict the ultimate use cases of the system, we want to test it on a variety of tasks and keep a set of capabilities in mind when designing the system. The basic principle here is that it should be able to solve any task the neocortex can solve. If we come up with a new mechanism that makes it fundamentally impossible to do something the neocortex can do, we need to rethink the mechanism. For example, thousand-brains systems should be able to model the world through any kind of movement-based sensory modality, from touch to echolocation. They should also be able to conceptualize abstract spaces, execute a series of intricate movements, and plan long-term actions. However, tasks such as multiplying arbitrary large numbers, or predicting the structure of a protein given its genetic sequence, are domains much better left to alternative technologies, such as calculators or deep-learning. 

The following is a list of capabilities that we always consider when designing and implementing the system. We are not looking for point solutions for each of these problems but a general algorithm that can solve all of them. It is by no means a comprehensive list, but it should give an idea of the scope of the system.

\begin{itemize}
  \item Recognizing objects independent of their location and orientation in the world.
  \item Determining the location and orientation of an object relative to the observer or to another object in the world.
  \item Recognizing an object and its pose by moving one sensor over the object.
  \item Performing flash inference (inference with no movement) by using many sensors in tandem.
  \item Performing learning and inference under noisy conditions. 
  \item Learning from a small number of samples.
  \item Learning from continuous interaction with the environment, maintaining previously learned representations.
  \item Learning without explicit supervision.
  \item Recognizing objects when other objects partially occlude them.
  \item Learning categories of objects and generalizing to new instances of a category.
  \item Learning and recognizing compositional objects, including novel combinations of their parts.
  \item Recognizing objects subject to novel deformations (e.g. Dali's `melting clocks', a crumpled up t-shirt, or recognizing objects learned in 3D but seen in 2D).
  \item Recognizing an object independent of its scale, and estimating its scale.
  \item Modeling and recognizing object states and behaviors (e.g. if a stapler is open or closed; whether a person is walking or running, and how their body evolves over time under these conditions).
  \item Using learned models to alter the world and achieve goals, including goals that require decomposition into simpler tasks.
  \item Generalizing modeling to abstract concepts derived from concrete models.
  \item Modeling language and associating it with grounded models of the world.
  \item Modeling other entities (Theory of Mind).
\end{itemize}

\section{Overview Of The Architecture}
\begin{figure}[h]
  \begin{center}
      \includegraphics[width=0.4\textwidth]{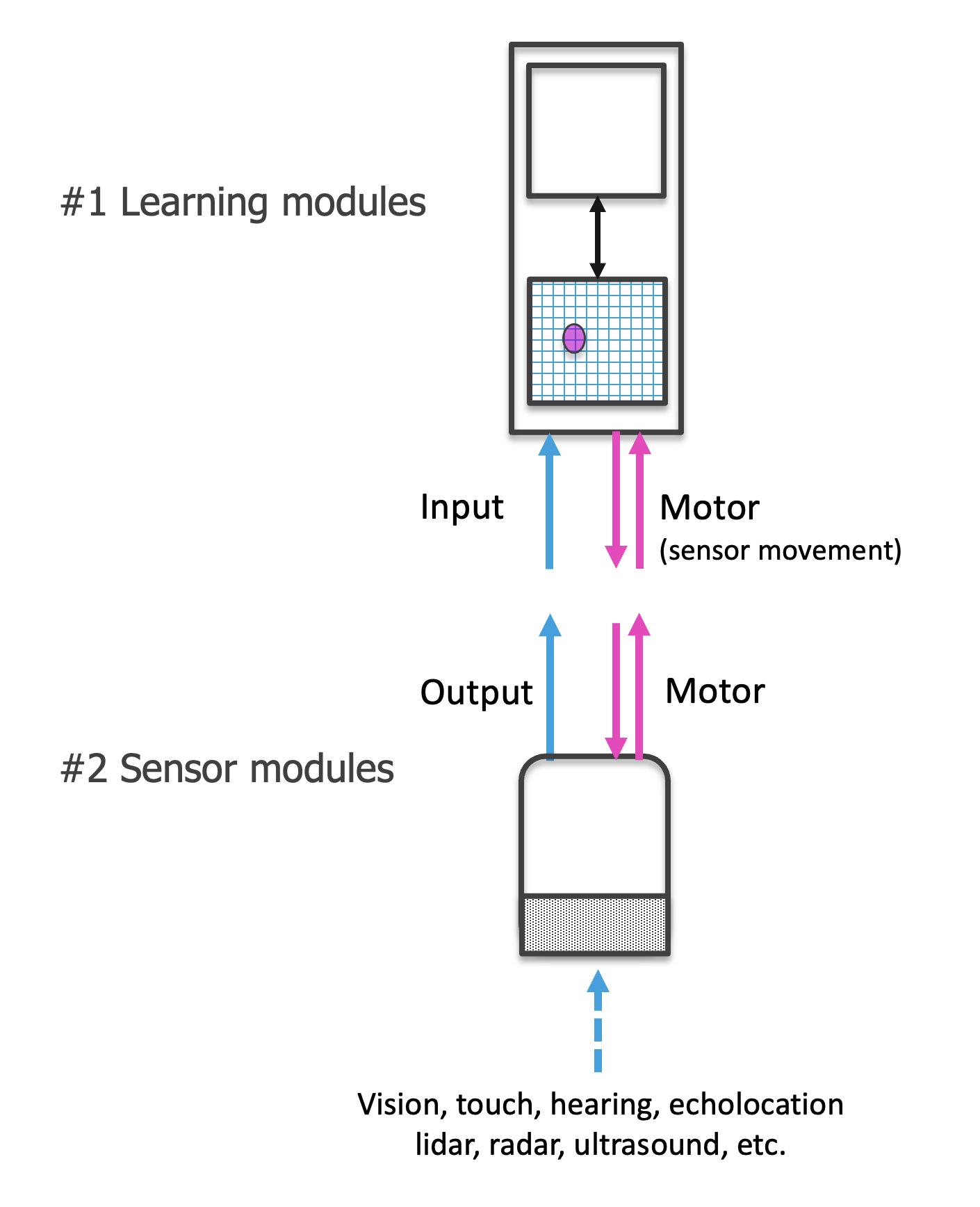}
  \end{center}
  \vspace{-10pt}
  \caption{Sensor modules receive and process the raw sensory and motor input. This is then communicated via a common messaging protocol to a learning module which uses this information to learn and recognize models of anything in the environment.}
  \label{fig:SMLM} 
\end{figure}

There are three major components that play a role in the architecture: sensor modules, learning modules, and the motor system. These three elements are tied together by a final key component, a common communication protocol. Due to this unified messaging protocol, the inner workings of each individual component can be quite varied as long as they have the appropriate interfaces. A simple example of a sensor module coupled to a learning module is shown in Figure \ref{fig:SMLM}, although we will begin by describing the CMP.

\subsection{Cortical Messaging Protocol}
\label{sec:CMP}

We use a common communication protocol that all components - learning modules, sensor modules, and the motor system - use to share information. By defining a consistent information format that sensor modules and learning modules must output, and that motor systems must receive, it is possible for all components to communicate with each other, and to combine them arbitrarily. Due to its inspiration from long-range connections in the cortex, we call this common communication protocol the Cortical Messaging Protocol (CMP)

In order to define the CMP, we must first define what we mean by an object, or a feature, in Monty. An \textit{object} is a discrete entity composed of a collection of one or more other objects, each with their own associated \textit{pose}. For example, an apple at a location and orientation in space is an object, but equally, an object could be a scene, an abstract arrangement of concepts, or any other composition of sub-objects. At the lowest level of this object hierarchy, an object is composed of inputs from a sensor module, which are also discrete entities with a location and orientation in space. Sensor inputs play a similar role to objects at other points in a hierarchy of learning modules, but the `objects' detected by sensors cannot be further decomposed. Wherever an object is being processed by a component of the system, it can also be referred to as a \textit{feature}. By convention, we usually refer to the input of a learning module as features and the output as an object ID. However, the object ID output of one learning module can become the feature input to the next learning module so they are by definition the same.

At its core, a CMP-compliant output contains \textit{a feature at a pose}. The pose contains a location in 3D space (naturally including 1D or 2D space) representing where the sensed feature is relative to the body, or another common reference point such as a landmark in the environment. In addition to location, the pose includes information about the feature's 3D orientation, which could be defined by the direction of a surface's point normal and its principal curvature, or the orientation of an object. Importantly, the message \textit{may} contain additional feature information, such as color, the magnitude of sensed curvature, or an object ID. Counter-intuitively, the nature of the feature does not need to be specified in the message for it to be a valid signal.

We highlight the choice that non-pose attributes of a feature are optional. This is contrary to many existing AI systems where models are often closer to bags-of-features and object structure is weakly represented, if at all. Here, the relative locations of features are more important than the features themselves. An example of how this aligns with human perception is how fruits arranged in the shape of a face can be easily recognized as a face, even though no typical face "features" are present. On the other hand, humans would not classify a random arrangement of eyes, a nose, and lips as a face.

Besides features and their poses, the standardized message also includes information about the sender's ID (e.g. the particular sensor module) and a confidence rating. Further below we discuss the internal models that learning-modules (LMs) develop - importantly, the CMP is never used to share these models between LMs. Instead, it can only communicate more abstract information about these models (such as an object ID).
The inputs and outputs of the system (raw sensory input to the SM and motor command outputs from motor modules) can have any format and do not adhere to any messaging protocol. They are specific to the agents sensors and actuators and represent the systems interface with the environment.Fin a common reference frame (e.g. relative to the body \footnote{In the following sections we may call this common reference frame \textit{"body-centric"}. In general, we just mean a common reference frame for all sensors. There may be applications without a concrete body (like several cameras set up in different locations of a room, a swarm of agents, or an agent navigating a more abstract space like the internet) where this just refers to an arbitrary point in space that all communicated poses are relative to.}). This makes it possible for all components to meaningfully interpret the pose information they receive. 

\subsection{Sensor Modules} 
Thousand-brains systems can work with any type of sensor (vision, touch, radar, LiDAR,...) and integrate information from multiple sensory modalities without effort. For this to work, sensor modules need to communicate information in a common language. Transforming raw sensory input into this common language is the job of the sensor module.

Each sensor module receives information from a small sensory patch as input. This is analogous to a small patch on the retina, or a patch of skin, or the pressure information at one whisker of a mouse. In the simplest architecture, one sensor module sends information to one learning module, which models this information. How such local sensory inputs are integrated across time and space will be covered in a moment when we discuss learning modules.

The information processing within the sensor module turns the raw information from the incoming sensor patch into the cortical messaging protocol (detailed in section \ref{sec:CMP}). This process can be compared to light hitting the retina and being converted into spikes, the output of biological neurons. Additionally, the pose of the feature relative to the body is calculated from the feature's pose relative to the sensor and the sensor's pose relative to the body. As such, each sensor module outputs the feature it senses, as well as the feature's pose (location and rotation) in body-centric coordinates. The availability of this pose information is central to how the thousand-brains architecture operates.

A general principle of the system is that any processing specific to a modality happens in the sensor module. The output of the sensor module is not modality-specific anymore and can be processed by any learning module. A crucial requirement here is that each sensor module knows the pose of the feature relative to the sensor. This means that sensors need to be able to detect features and poses of features. They also need to be able to keep track of their position in space. This could be directly provided from the system, inferred from sensory inputs (like optical flow), or calculated from efference copies of motor commands.

\subsection{Learning Modules}
\begin{figure}[h]
  \begin{center}
      \includegraphics[width=0.35\textwidth]{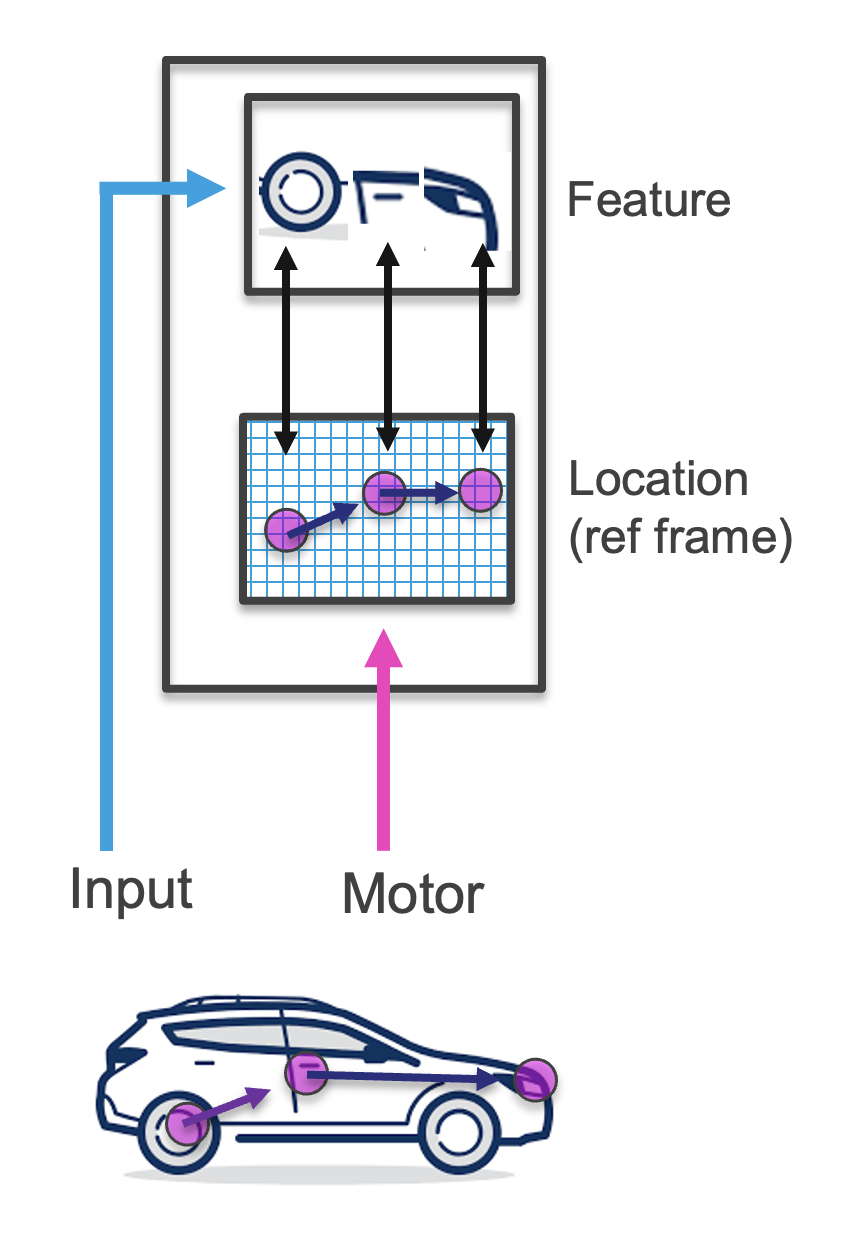}
  \end{center}
  \caption{Learning modules learn structured models through sensorimotor interaction, using reference frames. They model how incoming features are arranged relative to each other in space.}
  \label{fig:refframes}
\end{figure}

The basic building block for sensorimotor processing and modeling is the learning module (LM). These are repeating elements, each using the same input and output interface. Each LM should function as a stand-alone unit and be able to recognize objects on its own. Combining multiple LMs can speed up recognition (e.g. recognizing a cup using five fingers vs. one), allows for LMs to focus on storing only some objects, and enables learning compositional objects.

LMs receive features at poses. Features can either be feature IDs from a sensor module or object IDs (also interpreted as features) from a lower-level LM. The feature or object representation might be in the form of a discrete ID (e.g. the color red, a cylinder), or could be represented in a more high dimensional space (e.g. a vector of binary values representing hue, or corresponding to a fork-like object). Additionally, LMs receive the feature's or object's pose relative to the body, where the pose includes location and rotation. In this way, body-centric coordinates serve as a common reference frame for spatial computations, as opposed to the pose of features relative to each individual sensor. From this information, higher-level LMs can build up structured models of compositional objects (e.g. large objects or scenes).

The features and relative poses are incorporated into a model of the object. All models have an inductive bias towards learning objects within a 3-dimensional space, complimented by a temporal dimension. When interacting with the physical world, the 3D inductive bias is used to place features in internal models accordingly. However, the exact structure of space can potentially be learned, such that the lower-dimensional space of a melody, or the abstract space of a family tree, can be represented. 

\begin{figure*}[h]
  \begin{center}
      \includegraphics[width=0.7\textwidth]{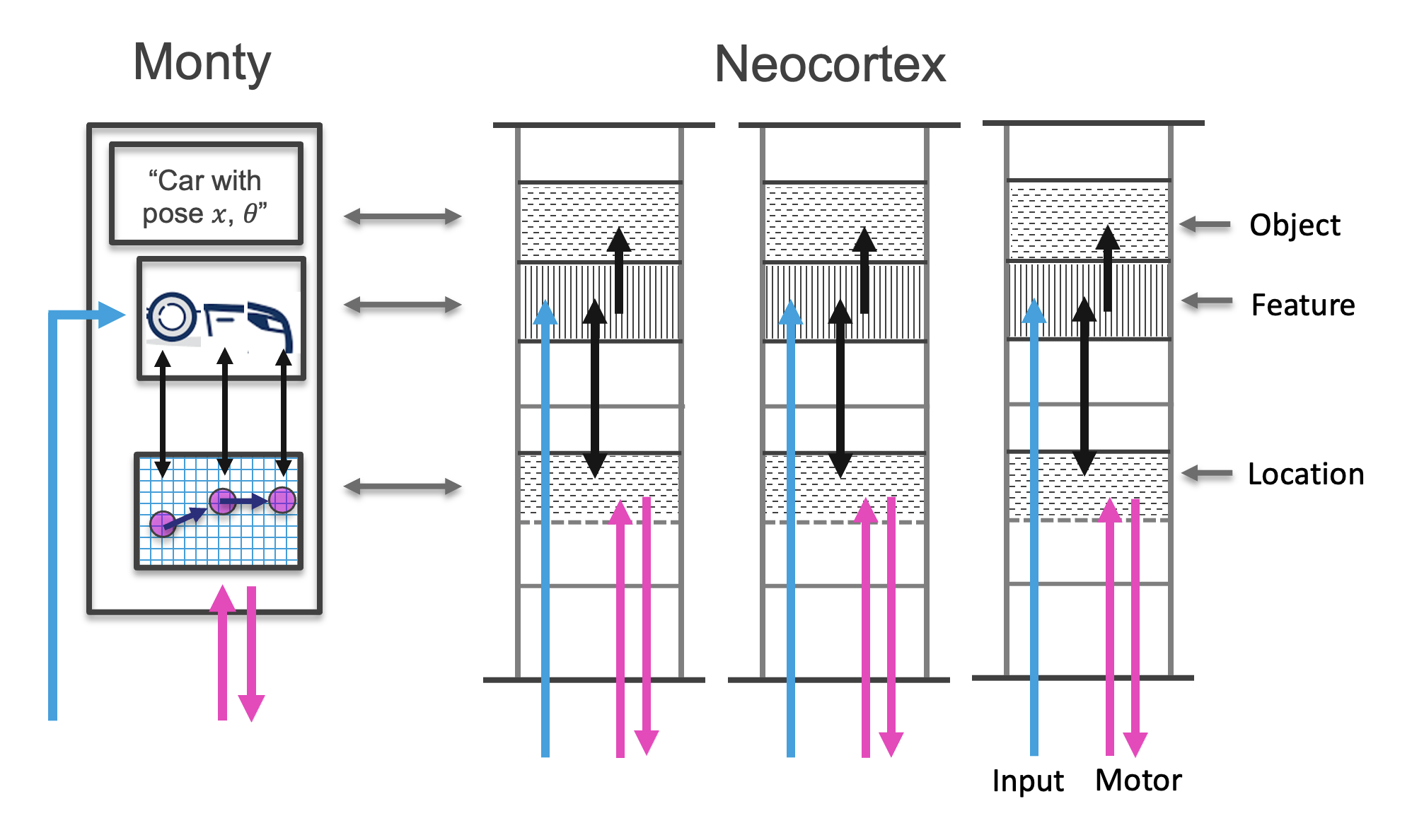}
  \end{center}
  \vspace{-10pt}
  \caption{Conceptual sketch of how the learning module could be implementing possible mechanisms of cortical columns. The figure on the right represents three cortical columns, including cellular layers. The internal structure of a learning module can be mapped onto these layers.}
  \label{fig:CCLM}
\end{figure*}

The LM, therefore, encompasses two major principles of the TBT: sensorimotor learning, and building models using reference frames (see Figure \ref{fig:refframes}). Both ideas are motivated by studies of cortical columns in the neocortex (see Figure \ref{fig:CCLM}), as well as \citet{Hawkins2017, Hawkins2019ANeocortex}.

Besides learning new models, the LM also tries to match the observed features and relative poses to already learned models stored in memory. Internally, LMs use displacements between consecutive poses and map them into the model's reference frame. This makes it possible to detect objects even at novel poses.

To generate the LM's \textit{output}, we need to get the pose of the sensed object relative to the body. We can calculate this from the current incoming pose (pose of the sensed feature relative to the body) and the poses stored in the model of the object. This pose of the object can then be passed hierarchically to another LM in the same format as the sensory input (features at a pose relative to the body where the feature is the inferred object ID). 

Once the LM has determined an object's ID and pose, it can use the most recent observations (and possibly collect more) to update its model. As such, LMs continually learn more about the world, and learning and inference are two closely intertwined processes.

\subsection{Motor Information and Action Policies}

Movement is central to how thousand-brains systems understand the world. The spatial nature of reference frames is dependent on integrating movement information so that a learning module knows where its input features are located at any given moment. The movement information (pose displacement) can be a copy of the selected action command (efference copy) or deduced from the sensory input. Without the efference copy, movement can be detected from information such as optical flow or proprioception. Sensor modules use movement information to update their pose relative to the body. LMs use it to update the hypothesized location of their incoming features within an object's reference frame.

While movement is clearly important for an LM to understand the outside world, it is also important that this movement is not random. What's more, an intelligent system should be able to exert influence on the external world. This is where policies become crucial.

Thousand-brains systems make use of a combination of model-free policies, corresponding to lower-level components of the system (sensor-module - motor-system loops), together with model-based policies based within LMs and using learned models to inform actions. 

Model-based policies use more computational resources to enable more principled movement, such as moving a sensor to a location that will minimize uncertainty about the currently observed object. These policies are derived from LMs, where each LM produces a motor output, analogous to the universal motor outputs found in cortical columns \citep{Prasad2020LayerTargets}. The motor output is formalized as a goal state and also adheres to the CMP. The goal state could, for example, use the learned models and current hypotheses to calculate a sensor state that resolves uncertainty about which of two possible objects is being observed. It can also help to guide directed and more efficient exploration of parts of objects that are currently underrepresented in the internal models. Different policies can be leveraged depending on whether we are trying to recognize an object or trying to learn new information about an object. Finally, policies can enable a learning module to change the state of the world, such as pushing a button, or changing the position of an object on a table. 

Hierarchy can also be leveraged for goal-states, where a more abstract goal-state in a high-level LM can be achieved by decomposing it into simpler goal-states for lower-level LMs. Importantly, the same LMs that learn models of objects are used to generate goal-states, enabling hierarchical, model-based policies.

Model-free policies are useful for purely sensory-based actions such as smoothly moving a sensor over the surface of an object, or attending to a prominent feature. Model-free policies can also learn to carry out frequently performed tasks in a dexterous and rapid manner, freeing computational resources required for model-based policies. Finally, goal-states generated by LMs must be transformed into motor commands for actuators - a process that recruits model-free policies (innate or learned) in the motor system.

In the brain, much of this processing occurs subcortically. In a thousand-brains system, this corresponds to the motor-system area. We note that the motor area does not know about models of objects that are learned in the LMs, and therefore needs to receive useful goal states from the LMs. These commands adhere to the CMP, but the outputs of the motor area will deviate from the protocol in order to interface with the actuators of the system. This means the motor system serves the reverse role of the sensor module, translating CMP-compliant goal states into the specific movement commands of the actuator it is connected to.

\subsection{Multi-LM Systems}

\begin{figure*}
  \begin{center}
      \includegraphics[width=0.9\textwidth]{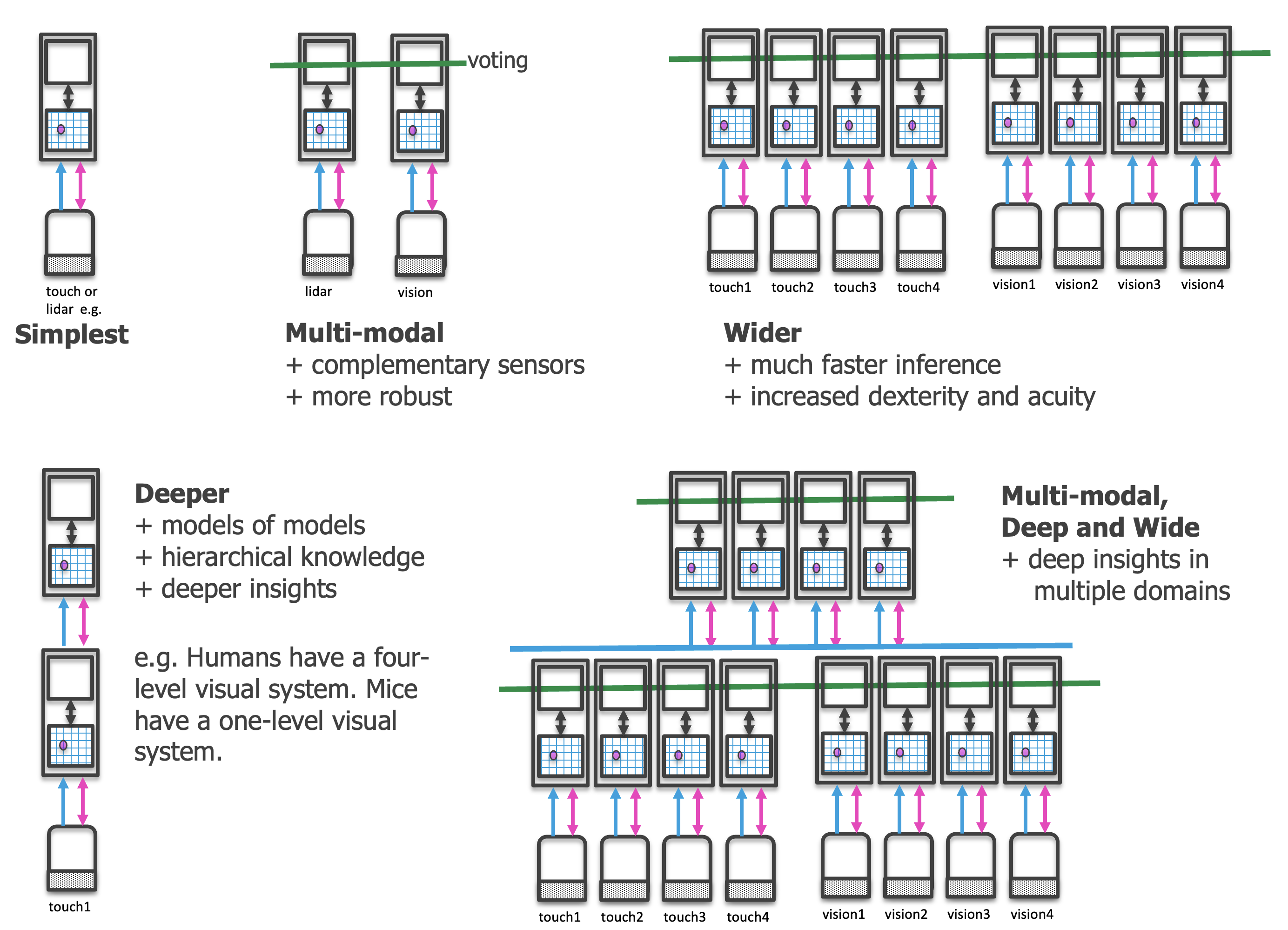}
  \end{center}
  \vspace{-10pt}
  \caption{By using a common messaging protocol between sensor modules and learning modules, the system can easily be scaled in multiple dimensions. This provides a straightforward way for dealing with multiple sensory modalities. Using multiple learning modules next to each other can improve robustness and speed through votes between them. Additionally, stacking learning modules on top of each other allows for more complex, hierarchical processing of inputs and modeling compositional objects.}
  \label{fig:Scaling}
\end{figure*}

Any given Monty system can be composed of multiple learning modules. Depending on their arrangement, LMs interact with one-another in a hierarchical manner, or via voting. A brief overview of these concepts is given below, while these possibilities are shown visually in Figure \ref{fig:Scaling}.

\subsubsection{Hierarchy: Composition and Learning on Different Spatial Scales} Learning modules can be stacked in a hierarchical fashion to process larger input patches and higher-level concepts. A higher-level LM receives feature and pose information from the output of a lower-level module and/or from a sensor patch with a larger receptive field, mirroring the connectivity of the cortex. The lower-level LM never sees the entire object it is modeling at once but infers it either through multiple consecutive movements and/or voting with other modules. The higher-level LM can then use the recognized model ID as a feature in its own models. This makes it more efficient to learn larger and more complex models as we do not need to represent all object details within one model. In particular, this enables the representation of \textit{compositional objects} by quickly associating different object parts with each other as relative features in a higher-level model. We discuss the importance of composition more later.

\subsubsection{Voting: Rapid Consensus} LMs share lateral connections in order to communicate their estimates of the current object ID and pose. This process, which we term \textit{voting}, adheres to the CMP, passing feature-pose information. Unlike connections between lower and higher LMs however, voting communicates a \textit{set} of all possible objects and poses under the current evidence (i.e. multiple messages adhering to the CMP). Through the lateral voting connections, LMs attempt to reach a consensus on which object they are sensing at the moment and its pose. This helps to recognize objects faster than a single module could.

We earlier highlighted that CMP messages are encoded in a common reference frame. This is key for voting to account for the relative displacement of sensors and, therefore, locations within LM models. For example, when two fingers touch a coffee mug in two different parts, one might sense the rim, while the other senses the handle. As such, `coffee mug' will be in both of their working hypotheses about the current object. When voting, they do not simply communicate `coffee mug', but also \textit{where} on the coffee mug other LMs should be sensing it, according to their relative displacements. As a result, voting is not simply a `bag-of-features' operation but is dependent on the relative arrangement of features in the world. 

Note that votes sent via the CMP do not contain any information about the input features received by that LM. For example, an LM might receive point-normals and surface curvature as its input features from an SM, and use this to model objects like coffee mugs and staplers. During voting, it will communicate its hypotheses around coffee mugs and staplers, but it will not communicate any information about sensed curvature to other learning modules. 

Finally, the CMP is independent of modality, and as such, LMs that have learned objects in different modalities (e.g. vision and touch), can still vote with each other to quickly reach a consensus. This voting process is inspired by the voting process described in \citet{Hawkins2017}.

\subsection{Bringing it Together}
\begin{figure*}
  \begin{center}
      \includegraphics[width=0.8\textwidth]{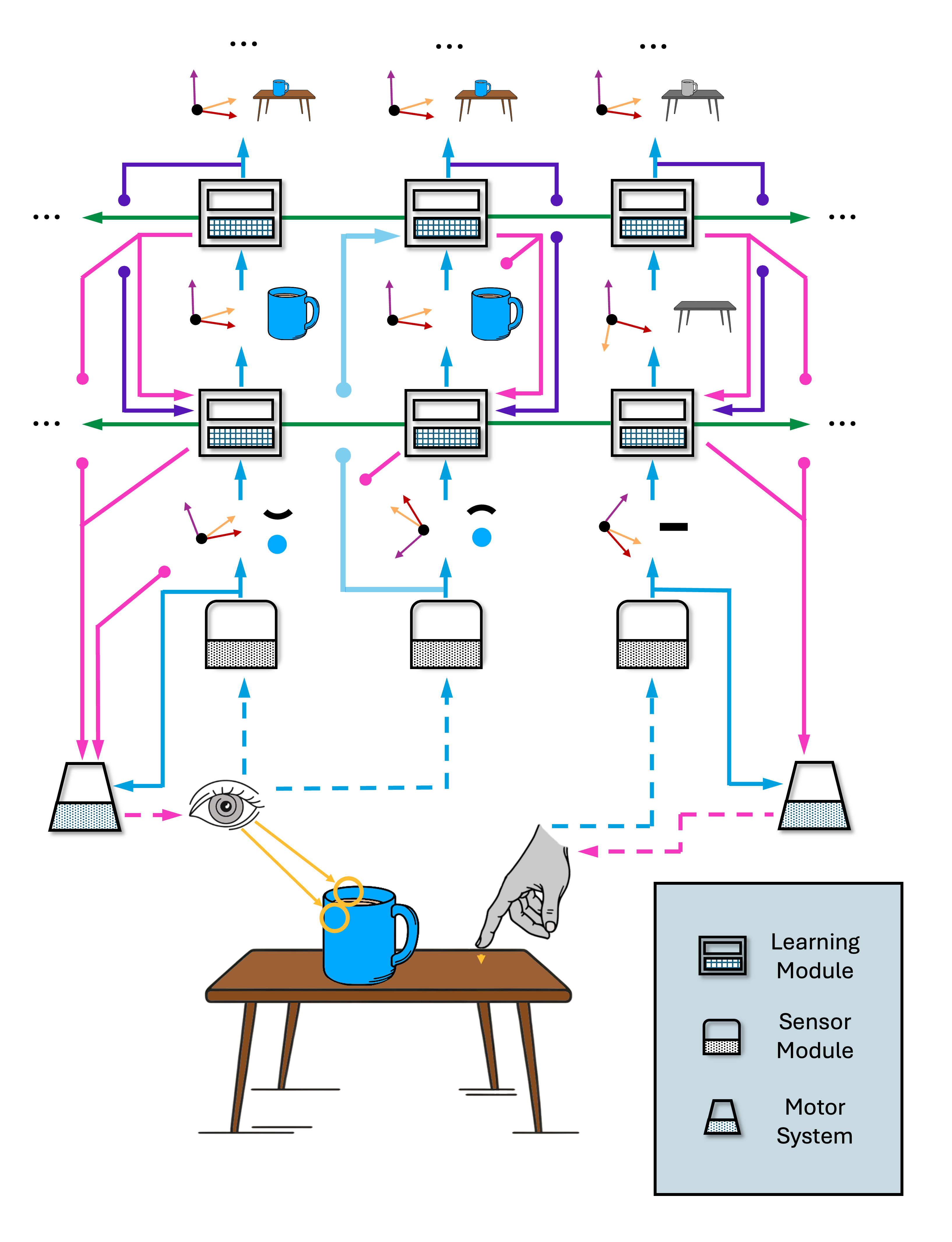}
  \end{center}
  \vspace{-10pt}
  \caption{High-level overview of the architecture with all the main conceptual components mirroring Figure \ref{fig:Scaling} applied to a concrete example. Blue lines indicate the feed-forward flow of information up the hierarchy. Purple lines show top-down connections, biasing the lower-level LMs. Green lines show lateral voting connections. Pink lines show the communication of goal states, which eventually translate into motor commands in the motor system. Every LM has a direct motor output. Information communicated along solid lines follows the CMP (contains features and pose). Discontinuations in the diagram are marked with dots on line ends. Dashed lines are the interface of the system with the world and do not need to follow the CMP. Blue dashed lines communicate raw sensory input from sensors. Pink dashed lines communicate motor commands to the actuators. The large, semi-transparent blue arrow is an example of a connection carrying sensory outputs from a larger receptive field directly to a higher-level LM.}
  \label{fig:overview}
\end{figure*}

To consolidate these concepts, please see Figure \ref{fig:overview} for an example instantiation of the system in a concrete setting. In this example, we see how the system could be applied to sensing and recognizing objects and scenes in a 3D environment using several different sensors, in this case, touch and vision. 

While provided to make the key concepts described above more concrete, bear in mind that this represents only one example of how the architecture can be instantiated. By design, thousand-brain systems can be applied to any application that involves sensing and active interaction with an environment. Indeed, this might include more abstract examples such as browsing the web, or interacting with the instruments that control a scientific experiment.

\section{Implementation}

\label{c:implementation}
We now describe in further detail the implementation of Monty, the first instance of a thousand-brains system. As just outlined in the general case, the basic components of Monty are: sensor modules (SM) to turn raw sensory data into a common language; learning modules (LM) to model incoming streams of data and use these models for interacting with the environment; motor system(s) to carry out actions, and translate abstract motor commands from the learning module into a format for actuators; and an environment in which the system is embedded and which it tries to model and interact with. The components within the Monty are connected by the Cortical Messaging Protocol (CMP) so that basic building blocks can be easily repeated and stacked on top of one another. Any communication within Monty is expressed as features at poses relative to a common reference frame such as the body. These CMP-compliant signals can be interpreted in different ways. For example, pose to the motor system is a target to move to, pose to an LM is the most likely sensed (from SM) or inferred (from LM) pose, and poses via voting connections are possible poses. 

All of these elements are implemented in Python at \url{https://github.com/thousandbrainsproject/tbp.monty} and their algorithmic details are described in the following sections. We begin by going into detail about general concepts, such as the experimental environment we currently employ, before describing the specifics of sensor modules, learning modules, and finally, action policies. This algorithm is under active development, and for a more detailed and the most up-to-date description, please refer to \href{https://thousandbrainsproject.readme.io/docs}{our documentation}.

Note that these descriptions refer to our current implementation and there will likely be many other implementations of the different components in the future. The idea of this framework is that any component can be customized and replaced as long as it follows the defined interface. For instance, one can switch out the type of learning module used without changing the sensor modules, environment, or motor system. Alternatively, one could implement a sensor module for a specific sensor and plug this into an existing learning module. Yet another possibility would be to test the current Monty configuration in another sensorimotor environment. The possibilities are endless, and the specific configuration and testbed described below is simply one instantiation that we found useful for designing this system.

\section{Experimental Evaluations}

The testbed currently used most often is focused on object recognition. While this involves learning models of objects and interacting with the environment, it all serves the purpose of recognizing objects and their poses. In the future, this focus will shift to settings where object pose and ID recognition subserves more complex interactions with the environment, such as manipulating the world to reach certain goal-states.

During an experiment, an agent collects a sequence of observations by interacting with an environment. We distinguish between training (internal models are updated using this sequence of observations) and evaluation (the agent only performs inference using already learned models but does not update them). In practice, Monty will always be learning, but establishing a distinct evaluation phase is useful for benchmarking performance in a controlled way.

For practical purposes, time is divided into discrete steps. We also divide an experiment into multiple episodes and epochs for easier measurement of performance. Overall, we discretize time in the three ways listed below. 

\begin{figure*}[t]
  \begin{center}
      \includegraphics[width=0.9\textwidth]{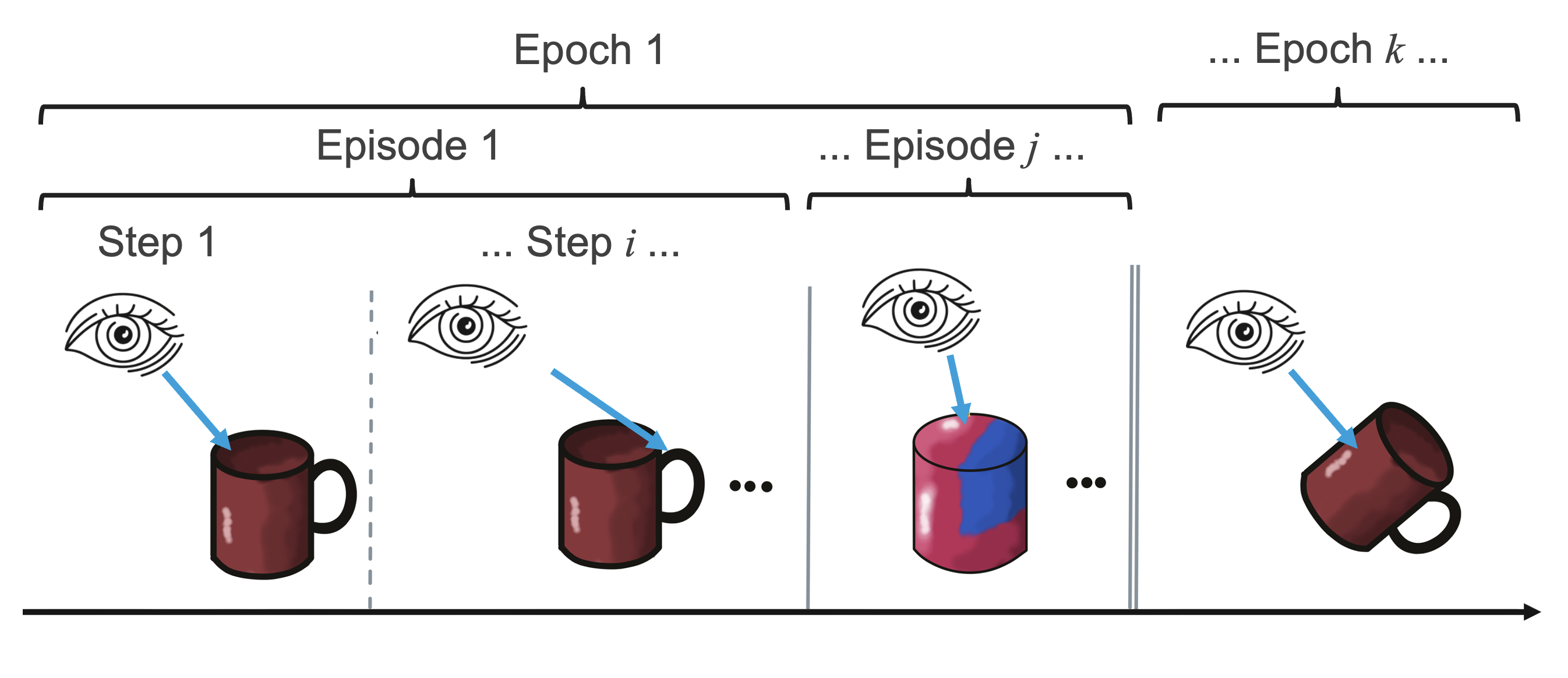}
  \end{center}
  \vspace{-10pt}
  \caption{Three ways time is discretized in our current experimental setup is into steps (one movement and one observation), episodes (take as many steps as needed to reach the terminal condition of the environment such as recognizing an object or completing a task), and epoch (cycle through all objects/scenarios in the dataset once). In this example, a red mug is observed in the first episode, and a two-tone painted cylinder is observed in the $j$th episode.}
  \label{fig:episodesepoch}
\end{figure*}

\begin{itemize}
  \setlength\itemsep{-0.1em}
  \item \textbf{Step:} Taking one action and receiving one observation. A step can happen at the level of Monty, as well as at the level of individual learning modules. The former includes steps that are deemed irrelevant for learning modules and where information is not sent to them, such as due to minimally changing inputs.  \item \textbf{Episode:} Putting a single object in the environment and taking steps until a terminal condition is reached, like recognizing the object or exceeding the maximum number of permitted steps.
  \item \textbf{Epoch:} Running one episode on each object/scene in the training or evaluation set of objects/scenes.
\end{itemize}

\section{Environment and Agent} 
\begin{figure*}
  \begin{center}
      \includegraphics[width=0.7\textwidth]{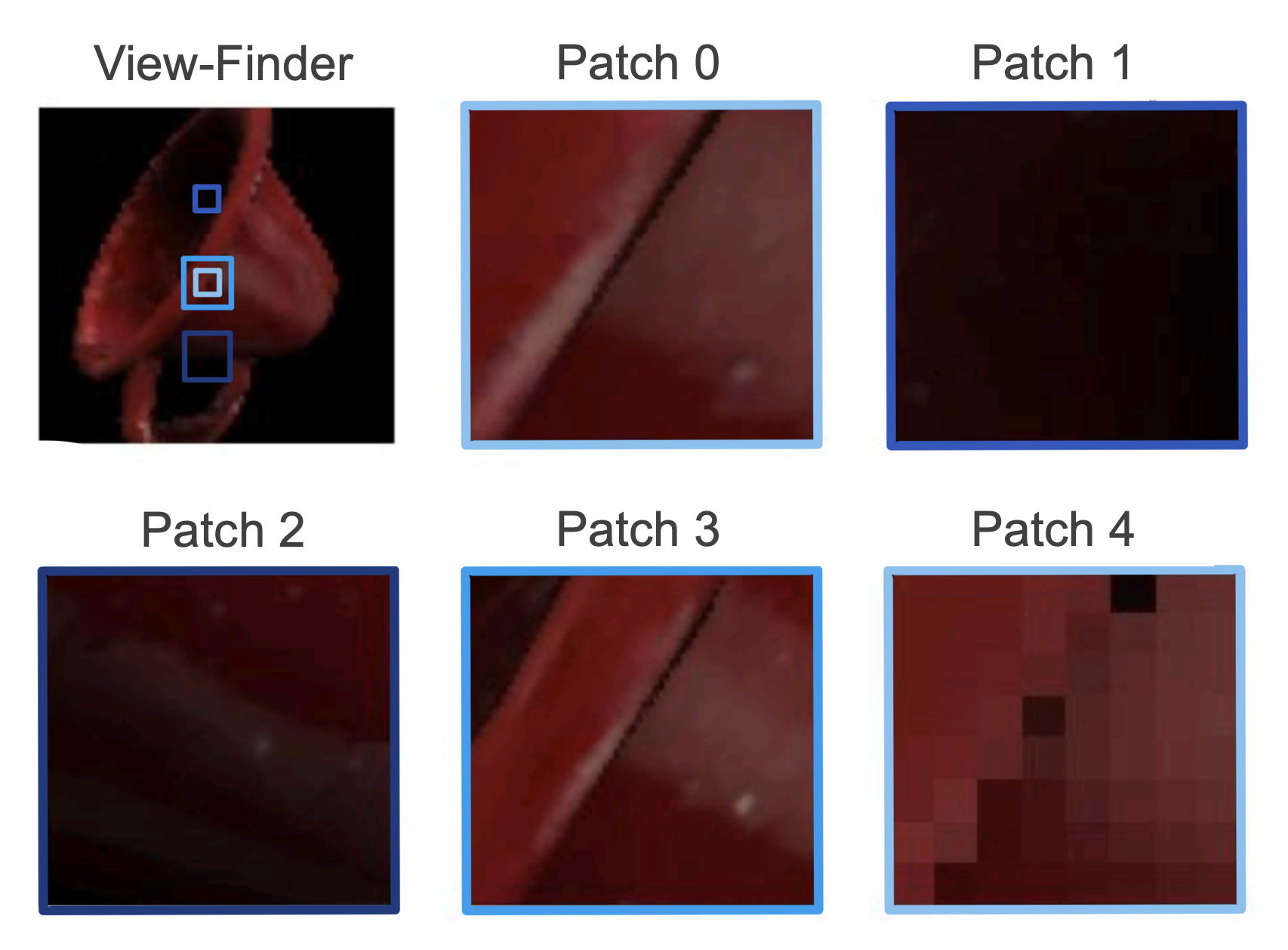}
  \end{center}
  \vspace{-10pt}
  \caption{Example of six sensors in Habitat. The view-finder is not connected to any sensor module or learning module and is only used to set up the experiment and for visualization. Each patch connects to one sensor module.}
  \label{fig:fivePatchExample}
\end{figure*}

The 3D environment and simulation engine used for most experiments is Habitat \citep{habitat19iccv, szot2021habitat, puig2023habitat3}. An agent is an actuator that can move independently in the environment, and has sensors coupled to it. Environments are currently initialized with one agent that has $N$ sensors attached to it. For most experiments, two sensors are used: the first sensor is the \textit{sensor patch} which is used for learning. It is a camera zoomed in 10x so that it can only perceive a small patch of the environment. The second sensor is a view-finder, which is at the same location as the patch and moves together with it, but its camera is not zoomed in. The view-finder is only used at the beginning of an episode to get a good view of the object and for visualization, but not for learning or inference (more details are in the discussion of policies found in Section \ref{sec:policies}). The agent setup can also be customized to use more than one sensor patch, such as the five patches in Figure \ref{fig:fivePatchExample}).  

\begin{figure*}
  \begin{center}
      \includegraphics[width=0.6\textwidth]{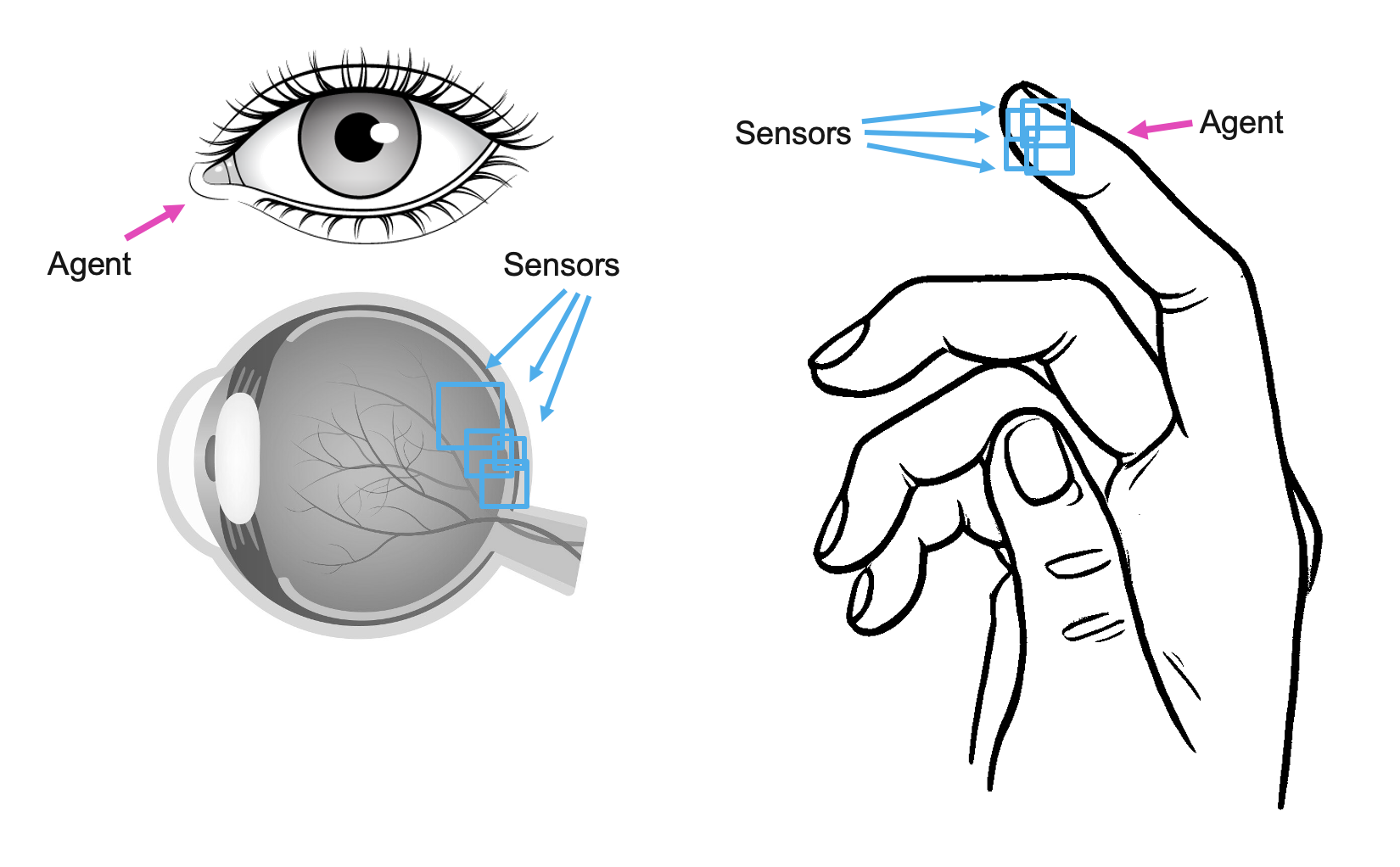}
  \end{center}
  \vspace{-10pt}
  \caption{Difference between sensors and agents. Agents can move independently of each other while all sensors connected to one agent move together. An agent itself does not perceive anything without sensors connected to it.}
  \label{fig:agentvssensor}
\end{figure*}

One can also initialize multiple agents (each with multiple sensors) and connect them to the same Monty instance. The difference between adding more agents vs. adding more sensors to the same agent is that all sensors connected to one agent move together, like neighboring patches on the retina. On the other hand, separate agents can move independently, like fingers on a hand (see Figure \ref{fig:agentvssensor}).

\begin{figure*}[t]
  \begin{center}
      \includegraphics[width=0.7\textwidth]{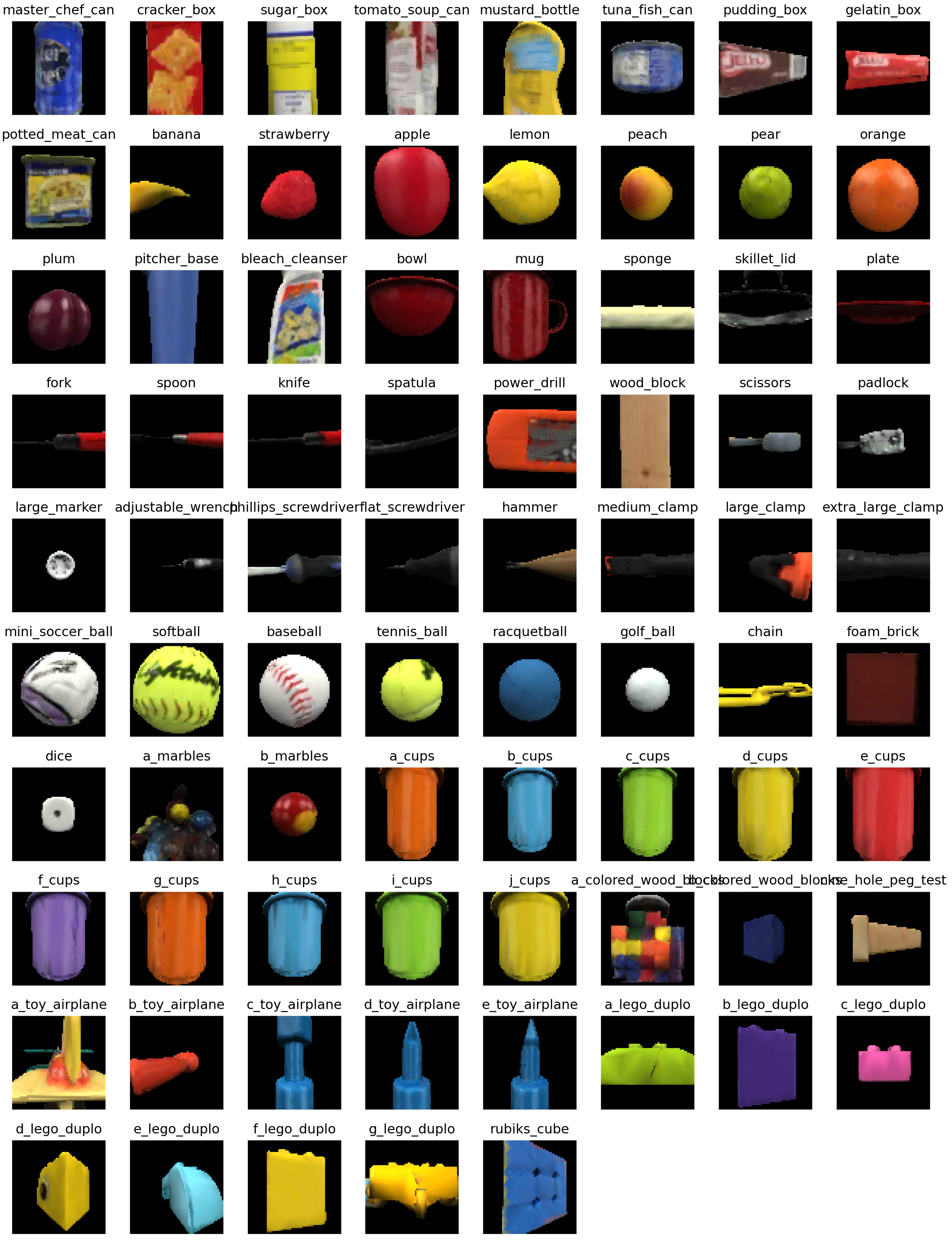}
  \end{center}
  \vspace{-10pt}
  \caption{The 77 objects of the YCB dataset at 0, 0, 0 rotation.}
  \label{fig:ycb}
\end{figure*}

The environment we typically evaluate object and pose detection in is an empty space with one object, although we are beginning to experiment with multiple objects. The object can be initialized in different rotations, positions and scales, although we do not currently vary the latter. For objects, one can either use the default Habitat objects (cube, sphere, capsule, etc.) or the YCB object dataset \citep{YCB}, containing 77 more complex objects such as a cup, bowl, chain, or hammer, as shown in Figure \ref{fig:ycb}. Currently there is no physics simulation so objects are not affected by gravity or touch and therefore do not move.

Of course, other datasets and objects can be used, and indeed we are not limited to 3D environments. For example, one data configuration lets agent move in 2D over images from the Omniglot dataset or photos from an RGBD camera. The only crucial requirement is that we can use an action to retrieve a new, action-dependent, observation from which we can extract a pose. Finally, we have implemented a simple dataset with physics-dependent objects that evolve over time, although we have not yet begun testing Monty in this setting. For a full list of current environments see \href{our documentation}{https://thousandbrainsproject.readme.io/docs/environment-agent}.

\section{The Monty Architecture}
The Monty architecture contains everything needed to model the environment and interact with it. It consists of sensor modules and learning modules, the communications wiring between them, and a motor system for carrying out actions.

Its specification consists of:

\begin{itemize}
  \setlength\itemsep{-0.1em}
  \item A list of SMs, each of which is responsible for processing raw sensory input and transforming it into a canonical format that any LM can operate on.
  \item A list of LMs, each of which is responsible for building models of objects given outputs from a sensor module.
  \item The mapping describing the precise coupling between SMs and LMs
  \item The mapping describing the precise coupling between LMs (for hierarchical and voting operations).
  \item A motor system responsible for moving the agent(s) of the system. This might also be implemented as motor modules, akin to sensor module.
  \item The mapping of sensors to an associated agent.
\end{itemize}

\begin{figure*}[t]
  \begin{center}
      \includegraphics[width=0.8\textwidth]{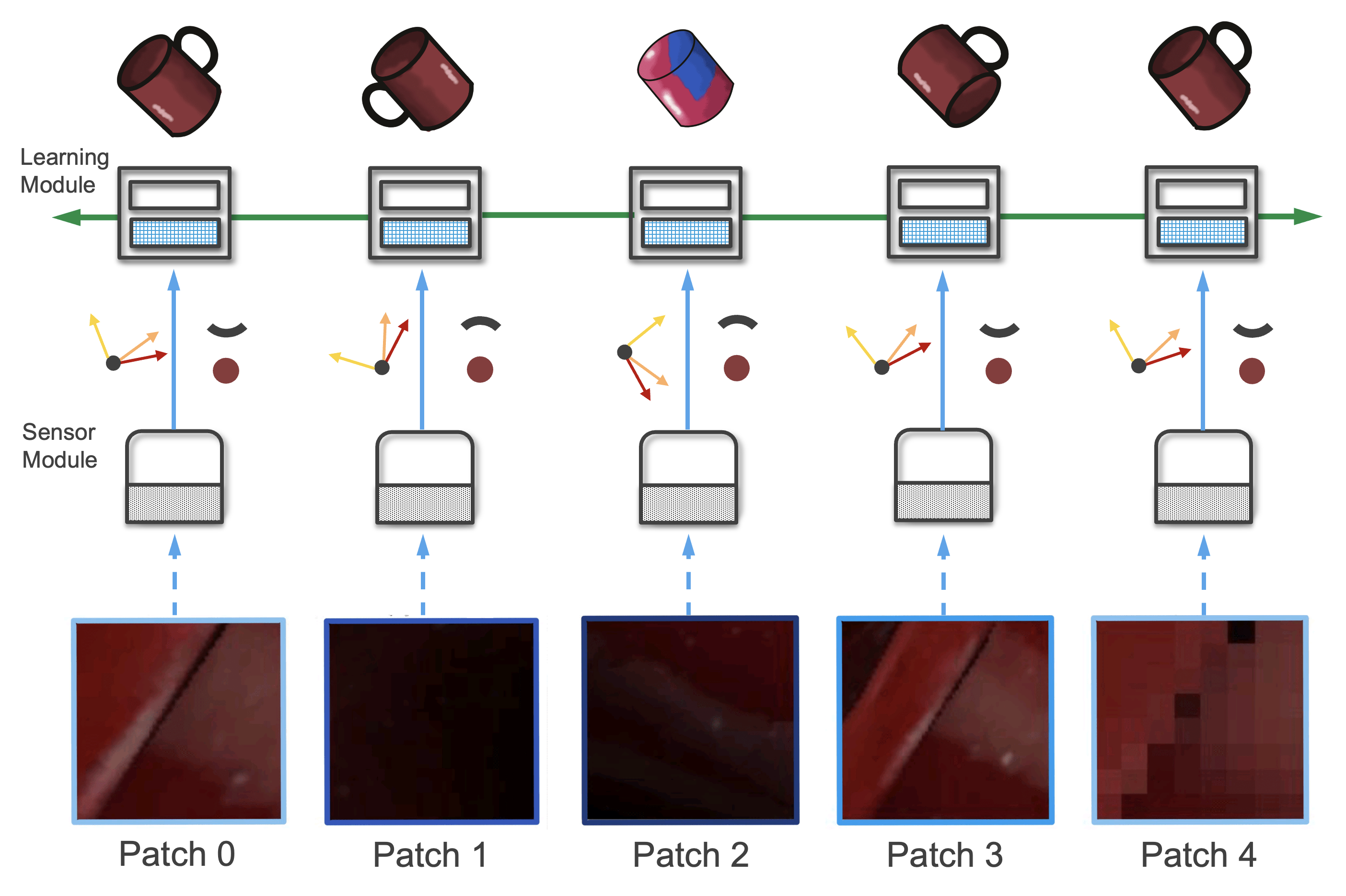}
  \end{center}
  \vspace{-10pt}
  \caption{Example Monty system with five sensor modules and learning modules. Each sensor patch perceives a small part of the environment, from which the associated SM extracts features and a pose (location and rotation relative to the body). This is sent to the LM which models the input and outputs another feature (most likely object ID) and its pose (most likely rotation and location of the object). LMs have lateral connections between one another (dotted lines) to communicate possible poses and narrow down their hypotheses faster. In this instance, most of the LMs believe the current object is a red mug, while the middle LM thinks it is most likely sensing a red-blue cylinder object that it has previously learned about. Through lateral voting the five LMs can quickly narrow down the possible object and pose.}
  \label{fig:fiveLMMonty}
\end{figure*}

Using the above information, we can specify the precise structure underlying an instance of Monty. For instance, if we have five sensors in the environment, we would specify five sensor modules, each corresponding to one sensor. Each sensor module could be connected to one learning module and the association between the learning modules is specified in an LM-LM connectivity matrix (specifying both lateral and hierarchical connectivity). This particular system's architecture would then look as shown in Figure \ref{fig:fiveLMMonty}.

\section{Observations and Sensor Modules}

The universal format that all sensor modules output is the CMP-compliant \textit{features at a pose} in 3D space. Each sensor connects to a sensor module which turns the raw sensory information into this format for down-stream processing. Each input to an LM therefore contains x, y, z coordinates of the feature location relative to the body, and three orthonormal vectors indicating its rotation. In sensor modules processing visual and tactile information, these pose-defining vectors are defined by the point normal and principal curvature directions sensed at the center of the patch. In learning modules (as detailed later), the pose vectors are defined by the detected rotation of an object. Additionally, the sensor module returns any sensed pose-independent features (e.g. color, texture, or curvature magnitude). The sensed features can be modality-specific (e.g., color for vision or temperature for touch), while the pose is modality-agnostic.

A CMP-compliant message must contain the following information:
\begin{itemize}
  \setlength\itemsep{-0.1em}
  \item Location (relative to the body or another common reference frame, such as a prominent feature in the environment)
  \item \textit{Morphological} features: 3x3 orthonormal vectors defining the orientation of the sensed feature.
  \item Non-morphological features: color, texture, curvature, etc.
  \item `Confidence' (defined in the range [0, 1]).
  \item A boolean for whether the message should be used (sent or processed downstream).
  \item Sender ID (a unique string identifying the sender) - this is used in an analogous way to the anatomical wiring found in the brain.
  \item Sender type (whether the sender is a sensor module or learning module).
\end{itemize}

A CMP-message is quite versatile, and depending on what system outputs it, it can be interpreted in different ways. Output by a sensor module, it can be seen as the observed features. When output by the learning module, it can be interpreted as the hypothesized or most likely object. As a motor output of an LM, it can be seen as a goal state (for instance, specifying the desired location and orientation of a sensor or object in the world). Lastly, when sent as lateral votes between LMs, it is interpreted as all possible objects and poses.

\begin{figure*}[t]
  \begin{center}
      \includegraphics[width=0.7\textwidth]{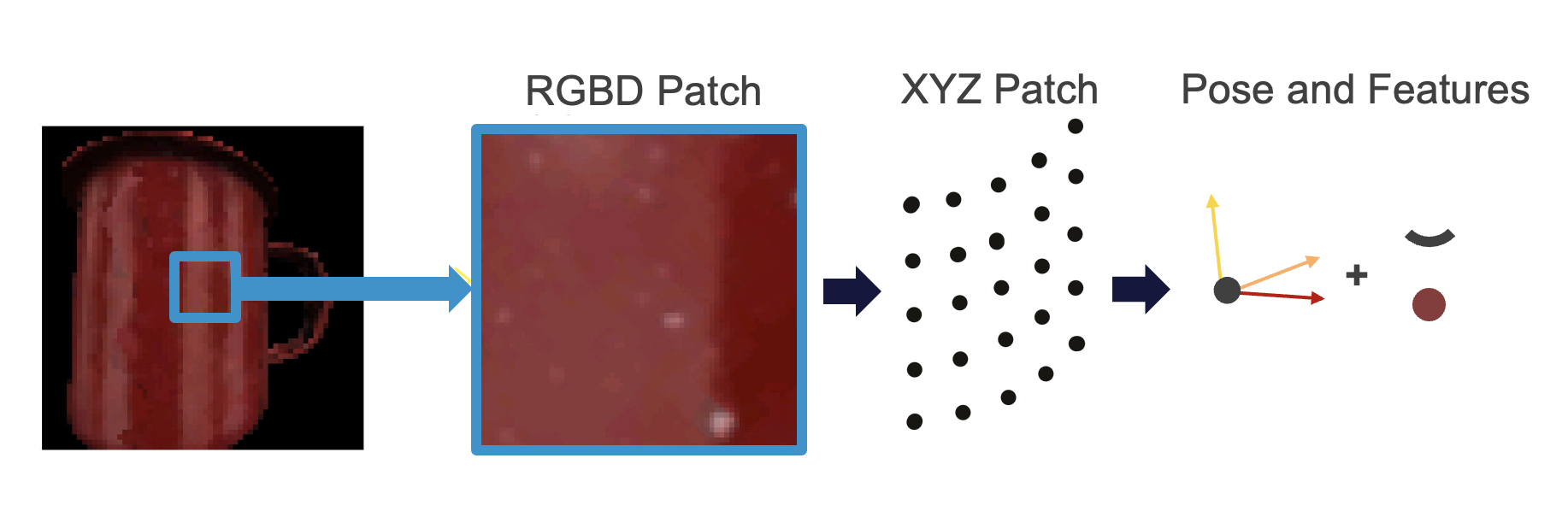}
  \end{center}
  \vspace{-10pt}
  \caption{Processing sensory inputs for the Cortical Messaging Protocol. The sensor patch captures a small area of the object (blue square), and if the sensor is a camera, it returns an RGBD image. We apply a transform to this image which calculates the x, y, z locations relative to the agent's body for each pixel using the depth values and the sensor location. From these points in space, the sensor module then calculates the point normal and principal curvature directions at the center point of the patch (pose). Additionally, the sensor module can extract pose-independent features such as color and the magnitude of curvature at the center of the patch. The pose (location + point normal and curvature direction) and features make up the observation at time step $t$ and are the output of the sensor module.}
  \label{fig:obsprocessing}
\end{figure*}

Note that some features are extracted using all of the information in a sensor patch (e.g. the locations of all points in the patch are used for point-normal and curvature calculation) but then refer to the center of the patch (e.g. only the curvature and point normal of the center are returned). At the moment, all the feature extraction is predefined, but in the future, some low-level feature extraction will likely be learned.

\section{Learning Modules}
Learning modules are the core modeling unit of Monty and are where all structured representations are learned. They are responsible for learning models from the incoming sensorimotor data, which they receive either from sensor modules, or other learning modules. Their input and output formats are features at a pose, and therefore comply with the CMP. Using the displacement between two consecutive inputs, they can learn object models of features relative to each other and recognize objects that they already know, independent of where they are in the world. How exactly this happens is determined by the details of the learning module - many possible architectures can be leveraged, although we will focus our description on the first generation of LMs that we have implemented.

Generally, each learning module contains a buffer, which functions as a short-term memory, and some form of long-term memory that stores models of objects. Both can then be used to generate hypotheses about what is currently being sensed, update, and communicate these hypotheses. Importantly, long-term memory relies on a structured representation of the world - a reference frame. At various times, such as when an object is recognized and being studied further, information from the buffer can be processed and integrated into the long-term memory. Finally, each learning module can also receive and send target states using a goal-state generator to guide the exploration and manipulation of the environment. We will discuss goal-state generators and model-based policies in more detail in a later section.

\subsection{Different Phases of Learning}

The learning module is designed to be able to learn objects from scratch. This means it is not assumed that we start with any previous knowledge or even complete objects stored in memory. As such, models in graph memory are updated after every episode, and learning and inference are tightly intertwined. If an object is recognized, the model of this object is updated with any newly sensed points. If no object is recognized, a new model is generated and stored in memory. This also means that the whole learning procedure can be unsupervised, as there are no object labels provided \footnote{Resetting the buffer at the end of an episode is a weak supervisory signal if we are changing the object after each episode, however this is not always the case, as different episodes often show the same object from different angles.}. 

To keep track of which objects were used for building a graph (since we do not provide object labels in this unsupervised learning setup), we store two lists in each learning module, mapping between learned graphs and the ground-truth objects observed in the world. These lists can later be used for analysis and to determine the performance of the system, but they are not used as a learning signal. This means learning can happen completely unsupervised without any labels being provided.

There are two modes the learning module could be in: \textit{training} and \textit{evaluation}. They are both very similar as both use the same procedure of moving and narrowing down the list of possible objects and poses. The only difference between the two is that in the training mode, the models in memory are updated after every episode. In practice, we often learn a series of objects and then save this Monty instance for a series of downstream evaluations with learning disabled. This enables controlled evaluations of the model, but it is important to emphasize that the long-term design of Monty is such that it would constantly be learning, with no separate learning and evaluation phases. 

The \textit{training mode} is split into two phases that alternate: The matching phase and the exploration phase. During the \textit{matching phase} the module tries to determine the object ID and pose from a series of observations and actions. This is the same as in evaluation. After a terminal condition is met (object recognized or no match found), the module goes into the \textit{exploration phase}. This phase continues to collect observations and adds them into the buffer the same way as during the previous phase; only the matching step is skipped. The exploration phase is used to add more information to the model representation at the end of an episode.

For example, the matching might terminate after three steps, telling us that the past three observations are not consistent with any models in memory. This would result in storing a new model in memory, however a model informed by only three observations is not very useful. Hence, we keep moving for a specified number of exploratory steps to collect more information about this object before adding it to memory. This is not necessary during evaluation since we do not update our models then.

\begin{figure*}[t]
  \begin{center}
      \includegraphics[width=0.7\textwidth]{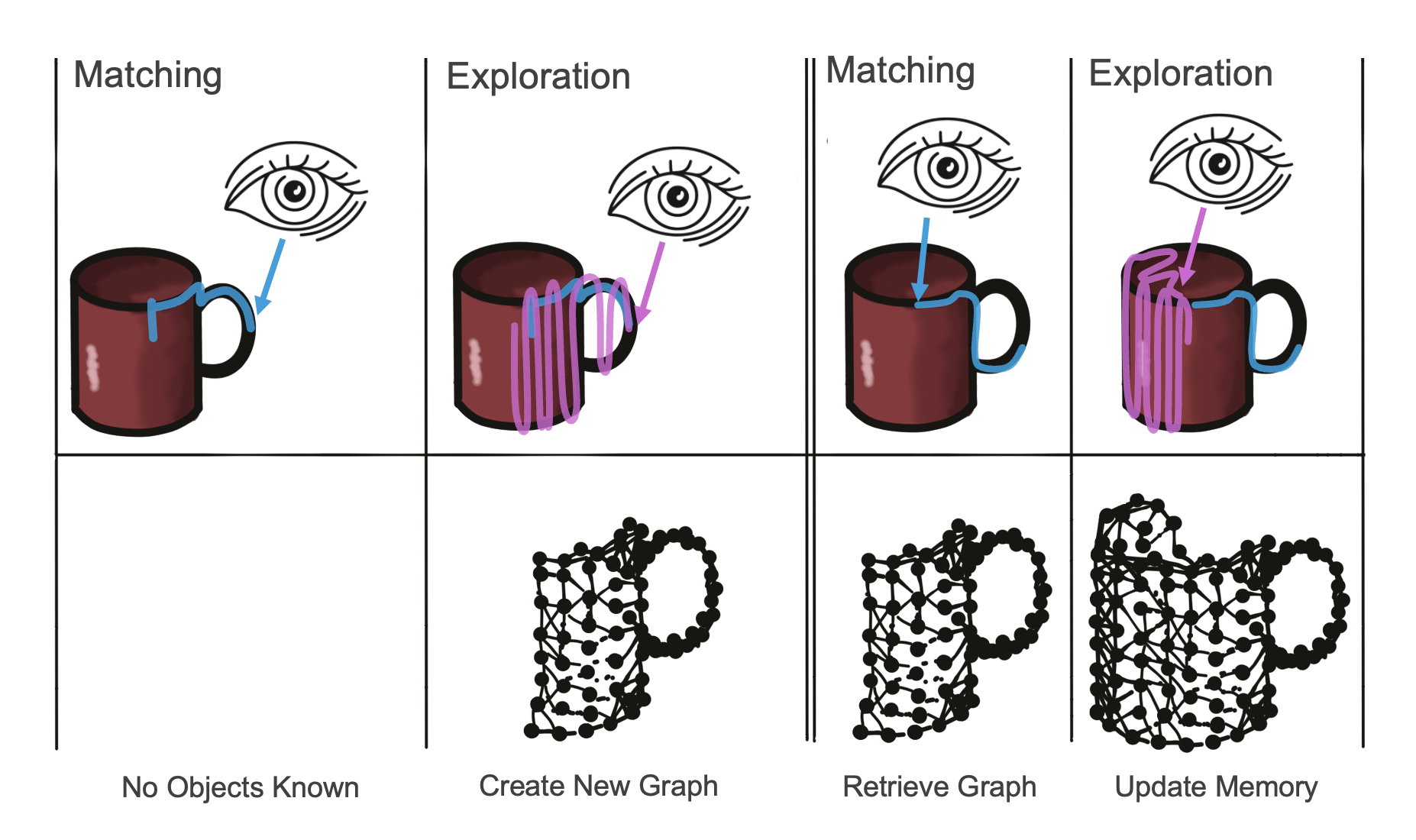}
  \end{center}
  \vspace{-10pt}
  \caption{First two episodes (separated by a vertical double line) during learning. After we recognize an object (matching phase, blue line) we can explore the object further to collect new information about it (exploration phase, pink line). This information can then be added to the model of the object in memory. The top row shows the agent's movements during the episodes. The bottom row shows the models in memory. As we are learning from scratch, we have no model in memory during the first episode.}
  \label{fig:learnfromscratch}
\end{figure*}

\subsection{First Generation Learning Modules}

\vspace{-6pt}

A learning module (LM) can take a variety of forms, as long as it models objects with reference frames, and interfaces via the Cortical Messaging Protocol (CMP). For example, an LM might use sparse-distributed representations (SDR) to represent features, and grid-cells for reference frames, using similar mechanisms as explored with synthetic objects in \citep{Lewis2019LocationsCells}. We have experimented with several variants, but the majority of our work has so far focused on LMs that leverage explicit, 3D graphs in Cartesian space. As such, these \textit{graph-based} LMs can be considered the first-generation of possible implementations. You may see occasional references to a `feature' or `displacement'-based graph-LM, however the \textit{evidence-based} LM is the implementation that we use as the default for all of our current experiments as it is most robust to noise and sampling new locations. As such, we will focus on describing its details. In brief, it makes use of graph-based reference frames where the evidence score associated with any node in the graph can be iteratively adjusted. 

We note that using explicit 3D graphs makes visualization more intuitive, improves interpretability, and facilitates debugging. This does not mean that we believe the brain stores explicit graphs with Cartesian coordinates. Future generations of LMs might use more neural representations such as grid-cells, however we are intentionally abstaining from such instantiations until they prove necessary.

\begin{figure}[t]
  \label{fig:graph}
  \begin{center}
      \includegraphics[width=0.4\textwidth]{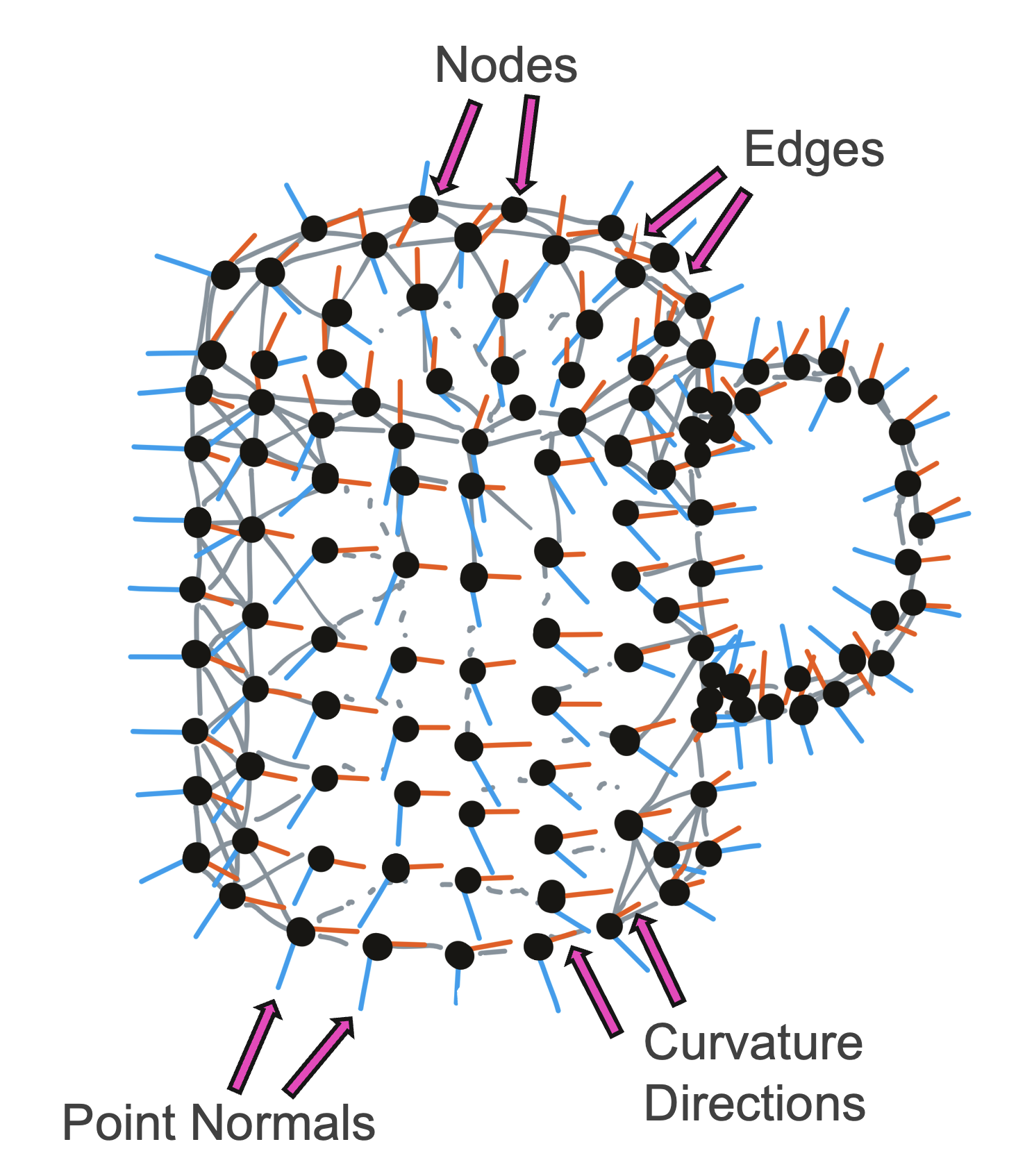}
  \end{center}
  \vspace{-10pt}
  \caption{A Graph of features (nodes), linked by displacements (edges). Each node represents a relative location and stores three pose vectors (for example, the point normal and the two principal curvature directions). Nodes can also have pose-independent features associated with them, such as color and curvature. The graph stored in memory can then be used to recognize objects from actual feature-pose observations.}
\end{figure}

\subsection{The Buffer (Short-Term Memory)}
Each learning module has a buffer that can be compared to short-term memory. The buffer only stores information from the current episode and is reset at the start of every new episode (and potentially at other events such as moving from one object onto another). Its content is used to update the graph memory at the end of an episode. The buffer is also used to retrieve the location observed in the previous step for calculating displacements.
\begin{figure*}[h]
  \begin{center}
      \includegraphics[width=\textwidth]{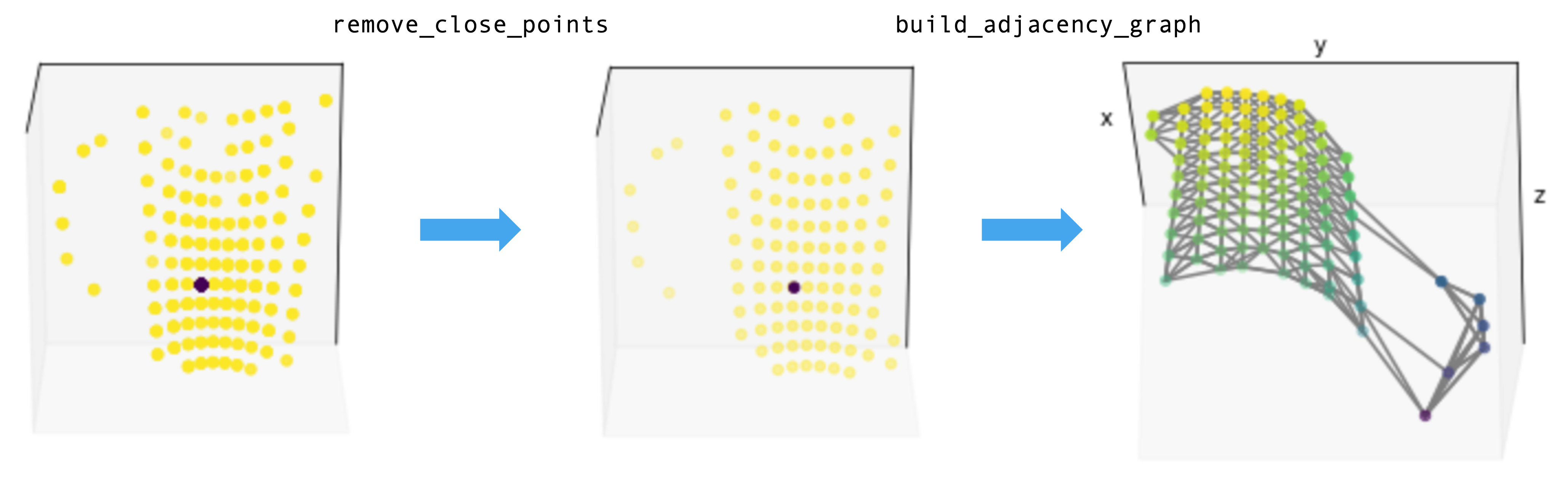}
      \includegraphics[width=\textwidth]{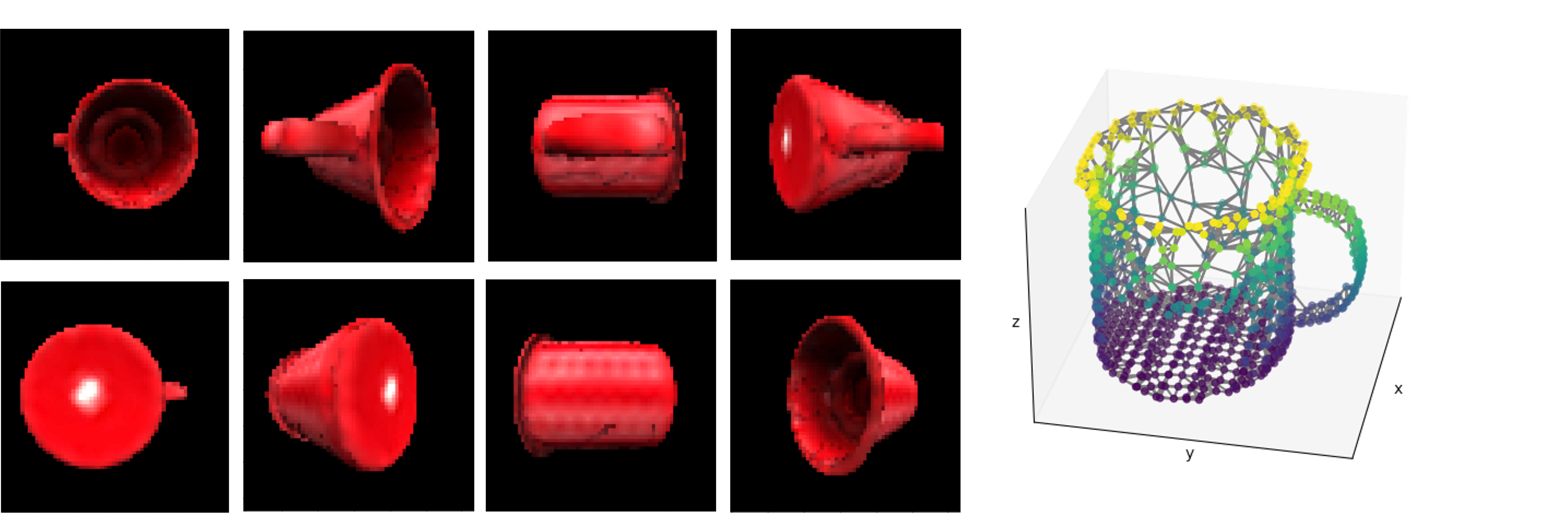}
  \end{center}
  \vspace{-10pt}
  \caption{(top) Building a graph from a buffer of observations. First, similar observations (high spatial proximity and feature similarity) are removed, and then the observations are turned into a graph structure as described above. (bottom-right) An object model that was learned from multiple views and extended over time.}
  \label{fig:multiviewmodel}
\end{figure*}

\subsection{The Graph Memory (Long-Term Memory)}
Each learning module has one graph memory which it uses as a long-term memory of previously acquired knowledge. In the graph learning modules, the memory stores explicit object models in the form of graphs in 3D Cartesian space. The graph memory is responsible for storing, updating, and retrieving models from memory.

\subsection{Object Models}
Object models are stored in the graph memory and contain information about one object. The information they store is encoded in reference frames and contains poses relative to each other and features at those poses. More specifically, the model encodes an object as a graph with nodes. Each node contains a pose and a list of features. Edge information can be used in principle (storing important displacements), but is not currently emphasized. Furthermore, graphs can generally be arbitrarily large in dimension and memory, although we are now experimenting with a form of constrained graphs that encourage intelligent use of limited representational capacity.

\subsection{Graph Building} 
\label{sec:graphbuilding}

A graph is constructed from a list of observations (poses, features). Each observation can become a node in the graph, which in turn connects to its neighbors in the graph by proximity or temporal sequence, indicated by the edges of the graph. Each edge has a displacement associated with it, which is the action that is required to move from one node to the other. Each node can have multiple features associated with it or simply indicate that there was information sensed at that point in space. Each node must contain location and orientation information in a common, object-centric reference frame.

\subsection{Graph Updates} 
\label{sec:graphupdates}
If a graph is not stored in memory yet, the LM will not find a match during object recognition, and it will add a new graph to memory.

Even if the object is already stored in memory, there may be new features we can learn about it and incorporate into the graph. For adding new observations, we need to know the pose of the object relative to our model in memory. The detected pose is used to rotate and translate new observations (which are relative to the sensor) into the reference frame of the object.

If a new point is too similar to those already in the graph by some threshold (such as being close in space or having similar features), then the LM will not add the point to its long-term memory. Avoiding the addition of similar points makes matching more efficient and avoids storing redundant information in memory. Instead, the LM stores more points where features change quickly (like where the handle attaches to the mug) and fewer points where features are not changing as much (such as on a flat surface).

\begin{figure*}[h]
  \begin{center}
      \includegraphics[width=0.7\textwidth]{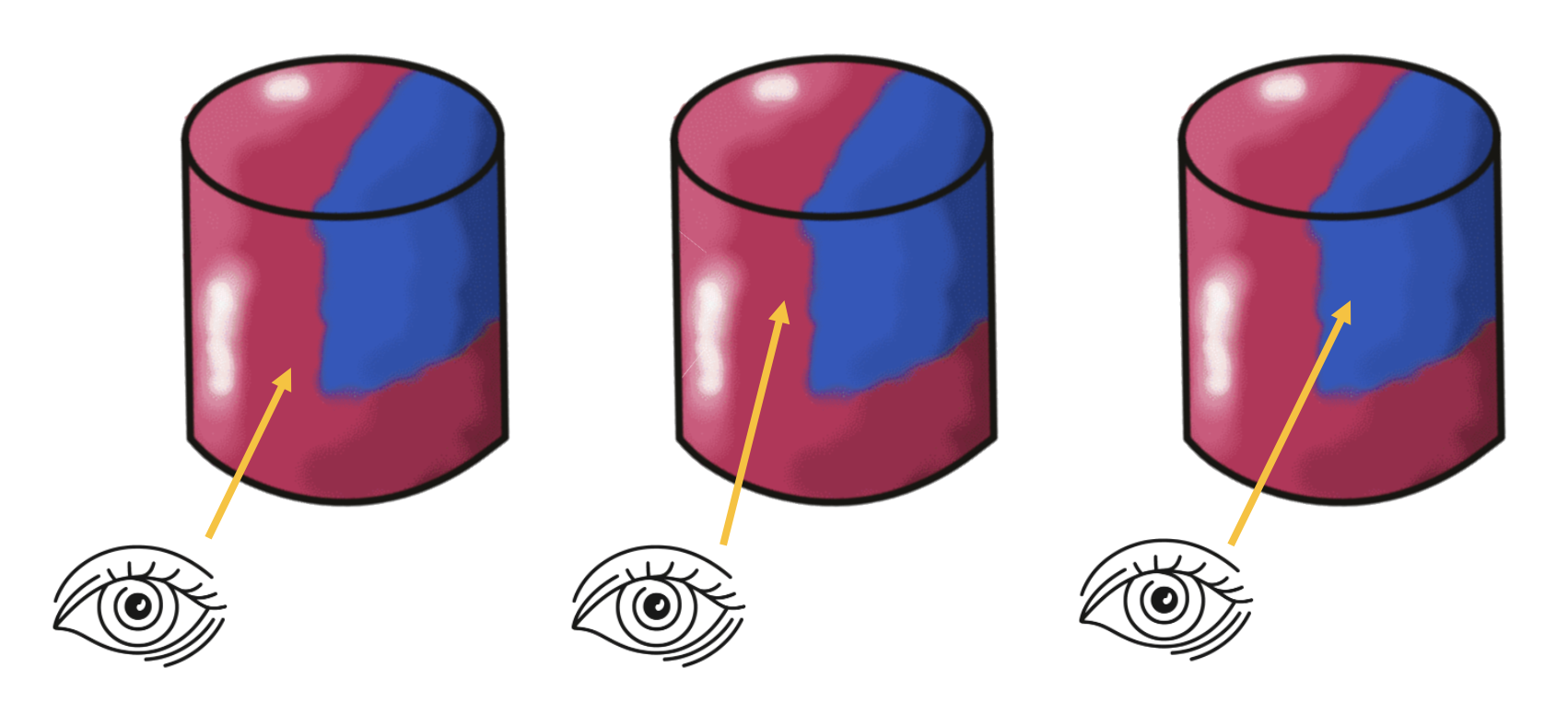}
  \end{center}
  \vspace{-10pt}
  \caption{\textbf{Sensing a Two-Toned Object} A representation of a sensor (camera patch), moving over the surface of a cylinder that is colored red in some parts, and blue in others. We will show how hypotheses are initialized and how evidence is updated based on this example. In this representation, the system collects three observations and performs two movements on the colored cylinder. The first observation is on the red rounded part of the cylinder (left), then it moves up (middle), and finally to the right onto the blue part (right).}
  \label{fig:Example}
\end{figure*}

\subsection{Using Graphs for Prediction and Querying Them} We can use graphs in memory to predict if there will be a feature sensed at the next location and what the next sensed feature will be, given an action/displacement (forward model). This prediction error can then be used for graph matching to update the possible matches and poses. 

A graph can also be queried to provide an action that leads from the current feature to a desired feature (inverse model). This can be used for a goal-conditioned action policy and more directed exploration. To do this, a hypothesis of the currently sensed object and its pose is required.

\subsection{The Evidence-Based Learning Module}
\label{sec:evLM}

The evidence-based LM uses a graph representation of objects, with all of the elements described up until now. In addition, a continuous evidence value is assigned to each hypothesis (which object and pose is being sensed), and these values are updated with every observation. We use the movement of a sensor on a colored cylinder, shown in figure \ref{fig:Example}, as an example.

\subsection{Initializing Hypotheses}
At the first step of an episode we need to initialize our hypothesis space. This means, we define which objects and poses are possible. 

\begin{figure*}[h]
  \begin{center}
      \includegraphics[width=0.7\textwidth]{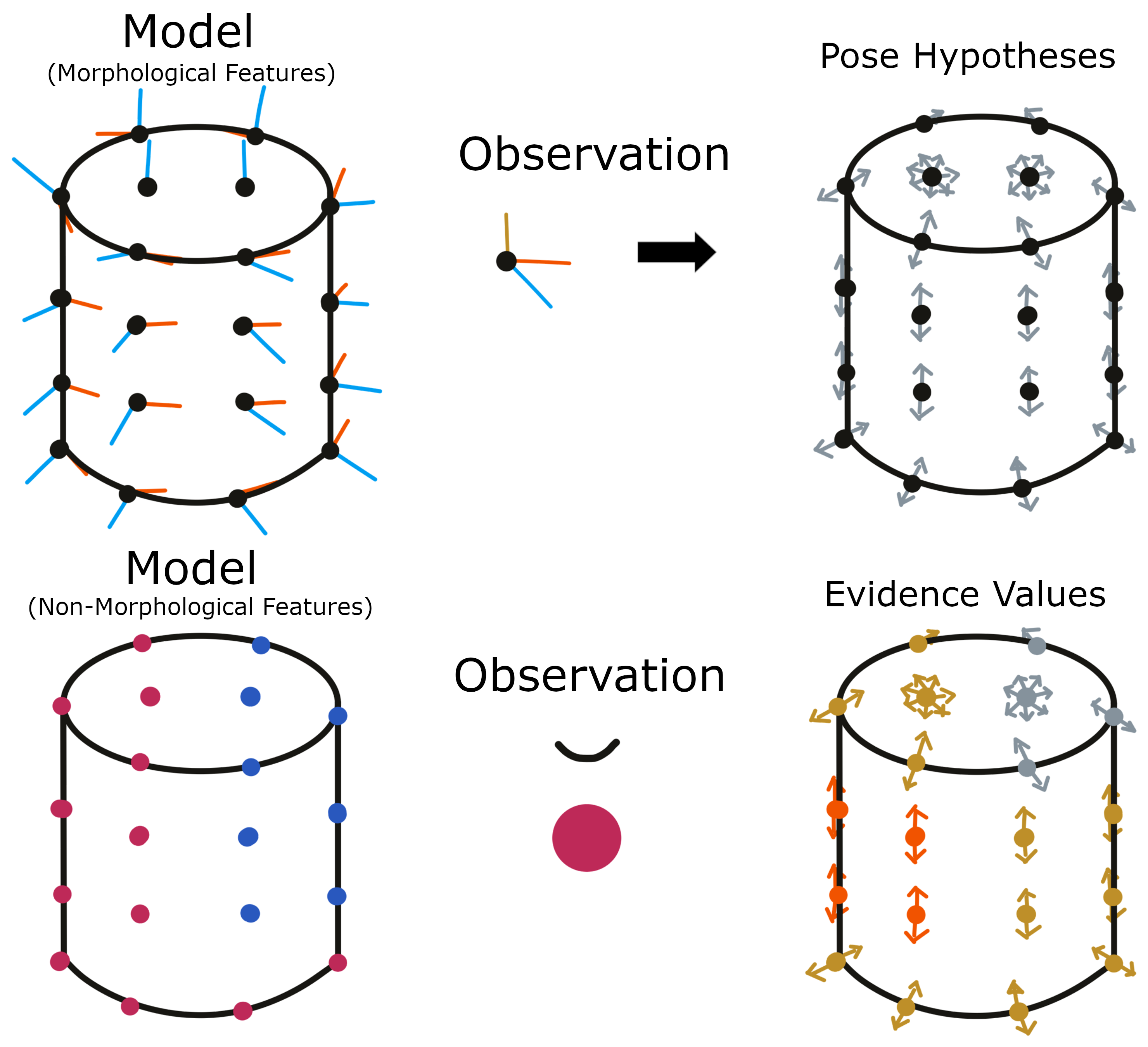}
  \end{center}
  \vspace{-10pt}
  \caption{Mechanism for initializing hypotheses from one observation. (Top row) Initializing possible poses. For this, we use all points in the stored model as possible locations and for each location, we calculate how we could rotate the sensed pose features such that they align with the stored pose features. (Bottom row) Initializing evidence counts for all hypotheses. If we do not sense any features, the evidence count is 0 for all possible poses, as signified in grey in the top right. We can then use sensed features and compare them to stored features at the locations in the model. If the difference is low, we add evidence proportional to that difference. The colors of dots in the model (bottom left) signify the stored color at that node. Colors in the hypothesis space (right) signify evidence where grey=0, yellow=medium, and red=high evidence. Note the four points that have high evidence because their stored features matched both the color and curvature of the observation.}
  \label{fig:InitHyp}
\end{figure*}

At the beginning of an episode, we consider all objects in an LM's memory as possible. We also consider any location on these objects as possible. We then use the sensed pose features (point normal and principal curvature direction) to determine the possible rotations of the object. This is done for each location on the object separately since we would have different hypotheses of the object orientation depending on where we assume we are. For example, the rotation hypothesis from a point on the top of the cylinder is 180 degrees different from a hypothesis on the bottom of the cylinder (see figure \ref{fig:InitHyp}, top).

By aligning the sensed point normal and curvature direction with the ones stored in the model we usually get two possible rotations for each possible location. We get two since the curvature direction has a 180-degree ambiguity, meaning we do not know if it points up or down as we do with the point normal. 

For some locations, we will have more than two possible rotations. This is the case when the first principal curvature (maximum curvature) is the same as the second principal curvature (minimum curvature), which happens, for example, when we are on a flat surface or a sphere. If this is the case, the curvature direction is meaningless and we sample N possible rotations along the axis of the point normal.

After initializing the hypothesis space we assign an evidence count to each hypothesis. Initially, this is 0 but if we are also observing pose-independent features such as color or the magnitude of curvature we can already say that some hypotheses are more likely than others (see figure \ref{fig:InitHyp}, bottom).

To calculate the evidence update, we take the difference between the sensed features and the stored features in the model. At any location in the model where this difference is smaller than the \textit{tolerance} value set for this feature, we add evidence to the associated hypotheses proportional to the difference. Generally, we never use features to subtract evidence, only to add evidence. Therefore, if the feature difference is larger than the tolerance (like in the blue and flat parts of the cylinder model in figure \ref{fig:InitHyp}) no additional evidence is added. The feature difference is also normalized such that we add a maximum of 1 to the evidence count if we have a perfect match and 0 evidence if the difference is larger than the set tolerance. A weighting factor associated with each type of feature can be used to emphasize some more than others.

In the example shown in figure \ref{fig:InitHyp} we initialize two pose hypotheses for each location stored in the model, except on the top and bottom of the cylinder where we have to sample more because of the undefined curvature directions. Using the sensed features, we update the evidence for each of these pose hypotheses. In locations where both color and curvature match, the evidence is the highest (red). In places where only one of those features matches, we have a medium-high evidence (yellow) and in areas where none of the features match we add 0 evidence (grey).

\subsection{Updating Evidence}
In all subsequent steps, we are able to use a pose \textit{displacement} for updating our hypotheses and their evidence. At the first step we had not moved yet so we could only use the sensed features. Now we can look at the difference between the current location and the previous location relative to the body and calculate the displacement. 

The relative displacement between two locations can then be used in the model's reference frame to test hypotheses. The displacement is only regarding the location while the rotation of the displacement will still be in the body's reference frame. To test hypotheses about different object rotations we have to rotate the displacement accordingly. We take each hypothesis location as a starting point and then rotated the displacement by the hypothesis rotation. The endpoint of the rotated displacement is the new possible location for this hypothesis. It basically says "If I would have been at location X on the object, the object is in orientation Y, and I move with displacement D, then I would now be at location Z". All of these locations and rotations are expressed in the object's reference frame.

\begin{figure*}[h]
  \begin{center}
      \includegraphics[width=0.95\textwidth]{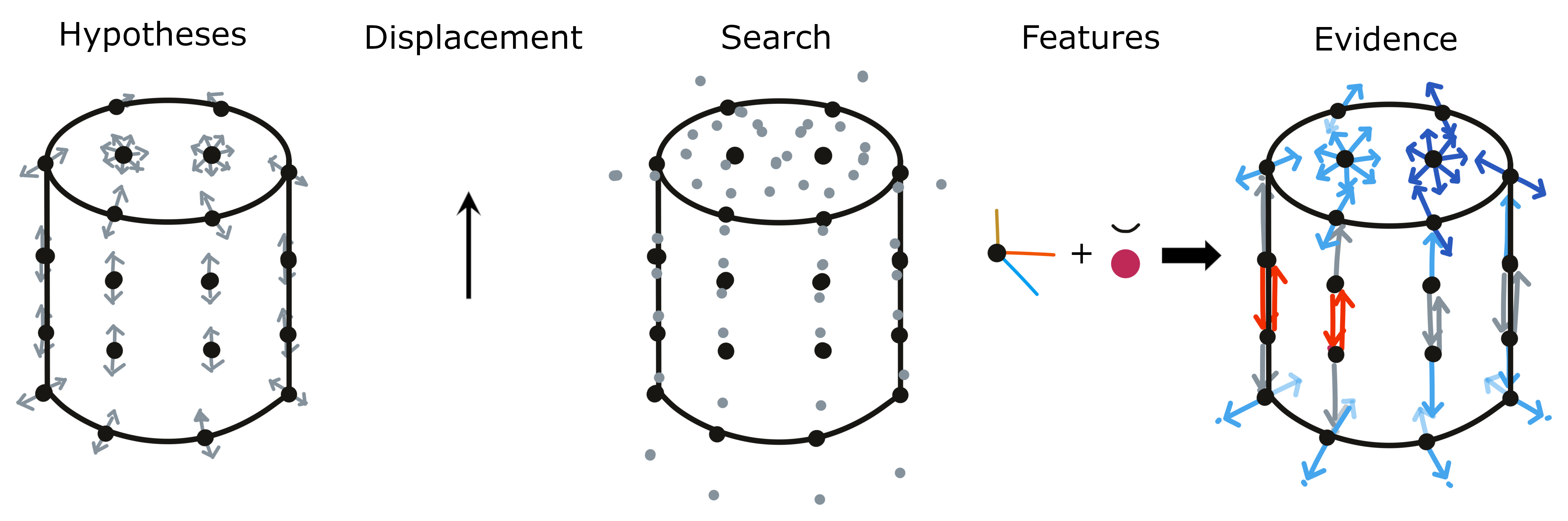}
      \includegraphics[width=\textwidth]{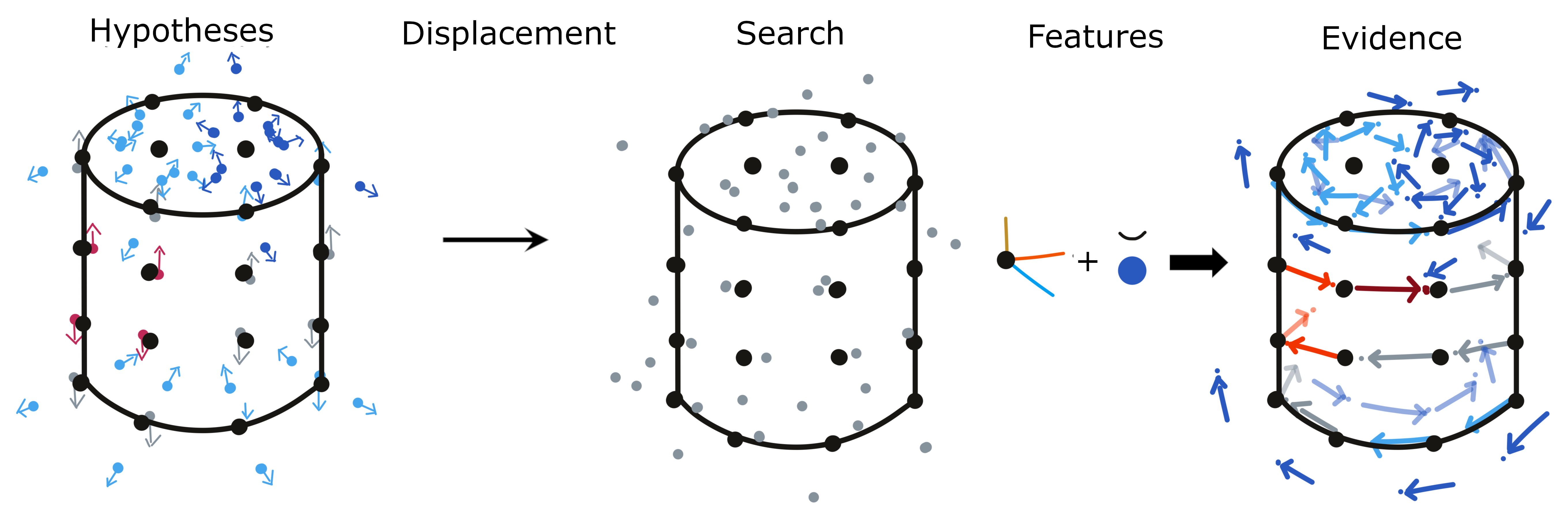}
  \end{center}
  \vspace{-10pt}
  \caption{Updating evidence using two steps. First, we move upwards and sense red and curved features (top row). Then, we move right and sense blue and curved features (bottom row). At each step, we take the hypotheses from the previous step (first column) and combine them with the sensed displacement (second column). Wherever these hypotheses combined with the displacement end up, defines our search locations (third column). We then look at the points stored in the model that are near each search location and compare the features stored there with the sensed features to update the evidence for each hypothesis (fourth and fifth column). Colors represent evidence values where dark blue=low, light blue=medium low, grey=0, orange=medium high, and red=high. Arrows in the left column signify pose hypotheses. Arrows in the right column signify the hypothesis plus the displacement and the updated evidence for this hypothesis.}
  \label{fig:evupdate}
\end{figure*}

Each of these new locations now needs to be checked, and the information stored in the model at this location needs to be compared to the sensed features. Since the model only stores discrete points, we often do not have an entry at the exact search location but look at the nearest neighbors.

\begin{figure*}[h]
  \begin{center}
      \includegraphics[width=0.85\textwidth]{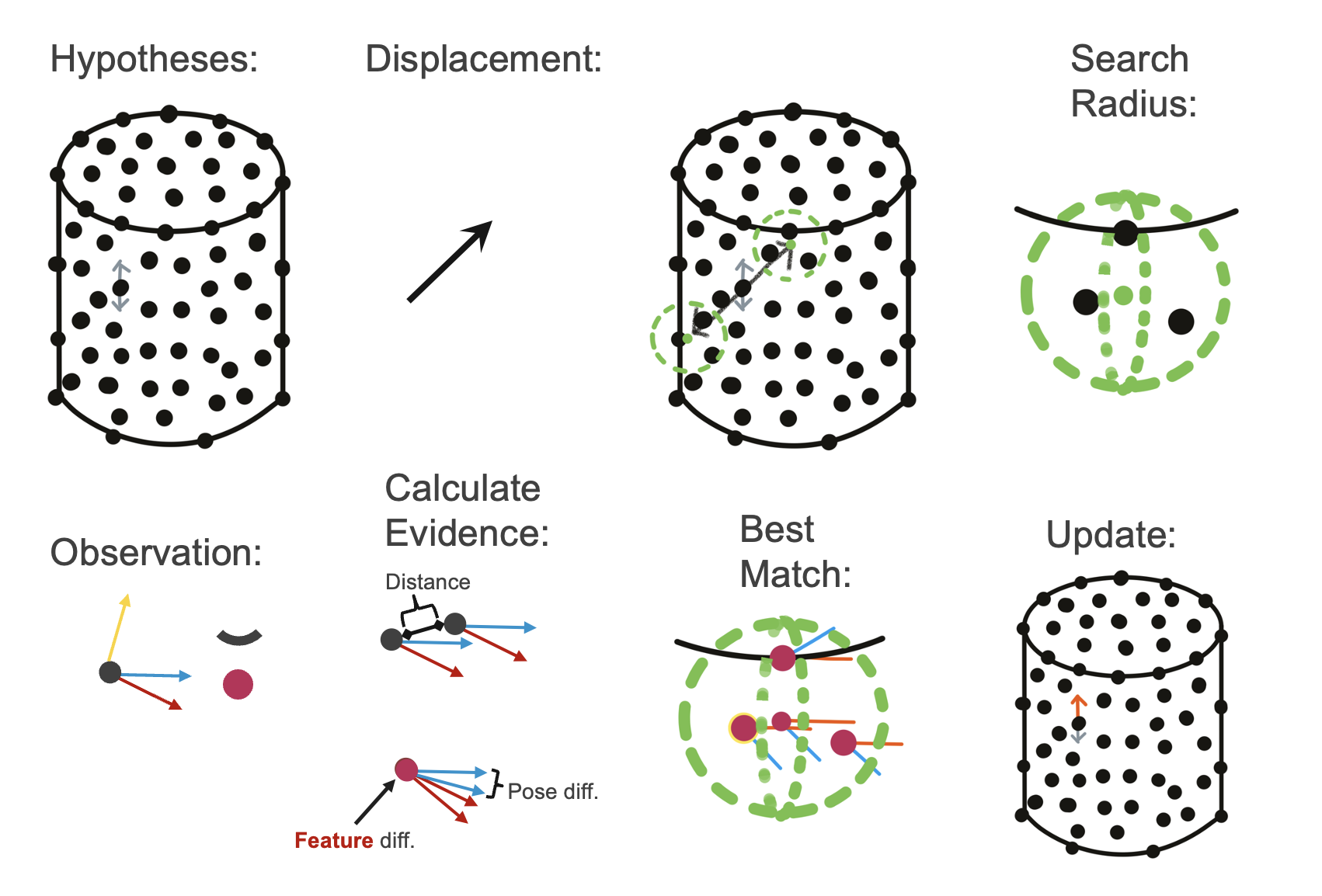}
  \end{center}
  \vspace{-10pt}
  \caption{Calculating the evidence update for one hypothesis. First, we calculate a search location given the hypothesis and displacement (top left). Then we find the nearest points stored in the model to this search location that are within a given radius (top right). For each point in the search radius, we compare the stored features to the sensed features and calculate the evidence (bottom left). We use the best match to update the hypothesis evidence (bottom right).}
  \label{fig:radius}
\end{figure*}

We now use both \textit{morphology} and \textit{features} to update the evidence. Morphology includes the distance of the search location to nearby points in the model and the difference between sensed and stored pose features (point normal and curvature direction). If there are no points stored in the model near the search location then our hypothesis is likely wrong and we subtract 1 from the evidence count. Otherwise, we calculate the angle between the sensed pose features and the ones stored at the nearby nodes. Depending on the magnitude of the angle we can get an evidence update between -1 and 1 where 1 is a perfect fit and -1 is a 180-degree angle (90 degrees for the curvature direction due to its symmetry). For the evidence update from the features, we use the same mechanism as during initialization where we calculate the difference between the sensed and stored features. This value can be between 0 and 1 which means at any step the evidence update for each hypothesis is in [-1, 2]. A detailed view of the nearest neighbor lookup and feature comparison to determine the evidence update for one of the hypotheses is shown in Figure \ref{fig:radius}.

The evidence value from this step is added to the previously accumulated evidence for each hypothesis. At the next step, the previous steps' search locations become the new location hypotheses. This means that the next displacement starts where the previous displacement ended given the hypothesis. We note that evidence updates are performed for all objects in memory. This can be done in parallel since the updates are independent of each other.

\begin{figure*}[h]
  \vspace{-10pt}
  \begin{center}
      \includegraphics[width=\textwidth]{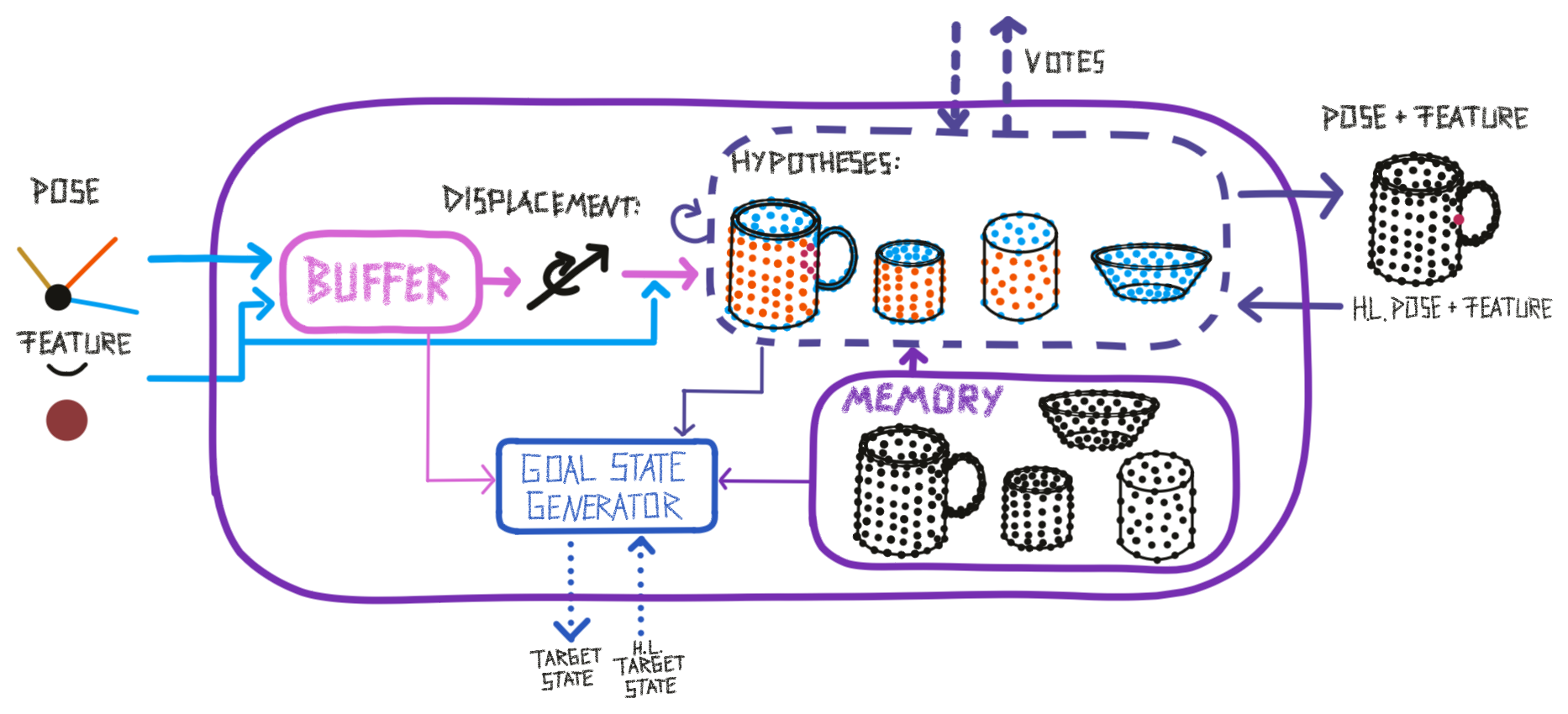}
  \end{center}
  \vspace{-10pt}
  \caption{Information flow in a graph learning module. At each step the LM receives features and a pose as input. Using the previous observation stored in the buffer it can calculate a pose displacement. This displacement, together with the sensed features, is used to evaluate all current hypotheses and update them. In the evidence LM this means updating their evidence, in the other two LMs it means to remove the hypotheses from the list of hypotheses if incoming information is inconsistent with the model's predictions. Using the current hypotheses and their evidence, the LM can then output a vote, which is sent to any connected LMs. If there are incoming votes, they will be used for another hypothesis update. Additionally, incoming top-down input can be used to modulate the evidence for different hypotheses. After this, the LM outputs its most likely hypothesis (object ID and pose) and a goal state (used for action selection). The goal state is produced by the goal-state generator, which can use a higher-level goal state and the LMs internal state and models to decide on the best goal state to output. Once matching is completed, the list of features and poses in the buffer can be used to update the graph memory.}
  \label{fig:learningM}
\end{figure*}

\subsection{Features and Morphology}
As mentioned before, features can only add evidence, not subtract it. Morphology (location and pose feature match) can add and subtract evidence. This is because we want to be able to recognize objects even when features are different. For example, if we have a model of a red coffee mug and are presented with a blue one we would still want to recognize a coffee mug. 

The idea is that features can add evidence to make recognition faster but they cannot reduce the likelihood of a hypothesis. This is only halfway achieved right now, since we consider relative evidence values and if features add evidence for some hypotheses and not for others, it also makes them implicitly less likely and can remove them from possible matches.

A future solution could be to store multiple possible features or a range of features at the nodes. Alternatively, we could separate object models more and have one model for morphology, which can be associated with many feature maps (kind of like UV maps in computer graphics). This is still an area of active conceptual development.

\section{LM Outputs and Connectivity}

A learning module can have three types of \textit{output} at every step, all of which adhere to the Cortical Messaging Protocol. The first one is, just like the primary, bottom-up input, a pose relative to the body and features at that pose. This could be the most likely object ID (represented as a feature) and its most likely pose. This output can be sent as input to another learning module or be read out to assess Monty's performance.

The second output is the LMs \textit{vote}. If the LM received input at the current step, it can send out its hypotheses and the likelihood of them to other LMs that it is connected to. For more details of how this works in the evidence LM, see section \ref{sec:evidencevoting}.

Finally, the LM can also suggest an action in the form of a \textit{goal state}. This goal state can then either be processed by another learning module and split into sub-goals or by the motor system and translated into a motor command in the environment. The goal state follows the CMP and therefore contains a pose relative to the body and features. For instance, the goal state might indicate a target pose for the sensor it connects to that would help it recognize the object faster, or would provide new information about an object for learning. More information on goal-states is provided in Section \ref{sec:policies}.

\subsection{Most Likely Hypothesis and Possible Matches}
We use continuous evidence values for our hypotheses, but for some outputs, statistics, and the terminal condition, we need to threshold them. This is done using a (currently user-set) percent-threshold parameter that defines how confident we need to be in a hypothesis to make it our final classification and move on to the next episode.

The threshold is applied in two places: To determine possible matches (the object we are observing) and possible poses (rotation of the object, and location on its surface). In either case, we look at the maximum evidence value and calculate the parameter-set percent of that value. Any object or pose that has evidence larger than the maximum evidence minus this relative percentage is considered possible. 

The larger the percent threshold is set, the more certain the model has to be in its hypothesis to reach a terminal state. This is because the terminal state checks if there is only one possible hypothesis. If we, for instance, set the threshold at 20\%, there cannot be another hypothesis with an evidence count above the most likely hypothesis evidence minus 20\%.

\begin{figure*}
  \begin{center}
      \includegraphics[width=0.8\textwidth]{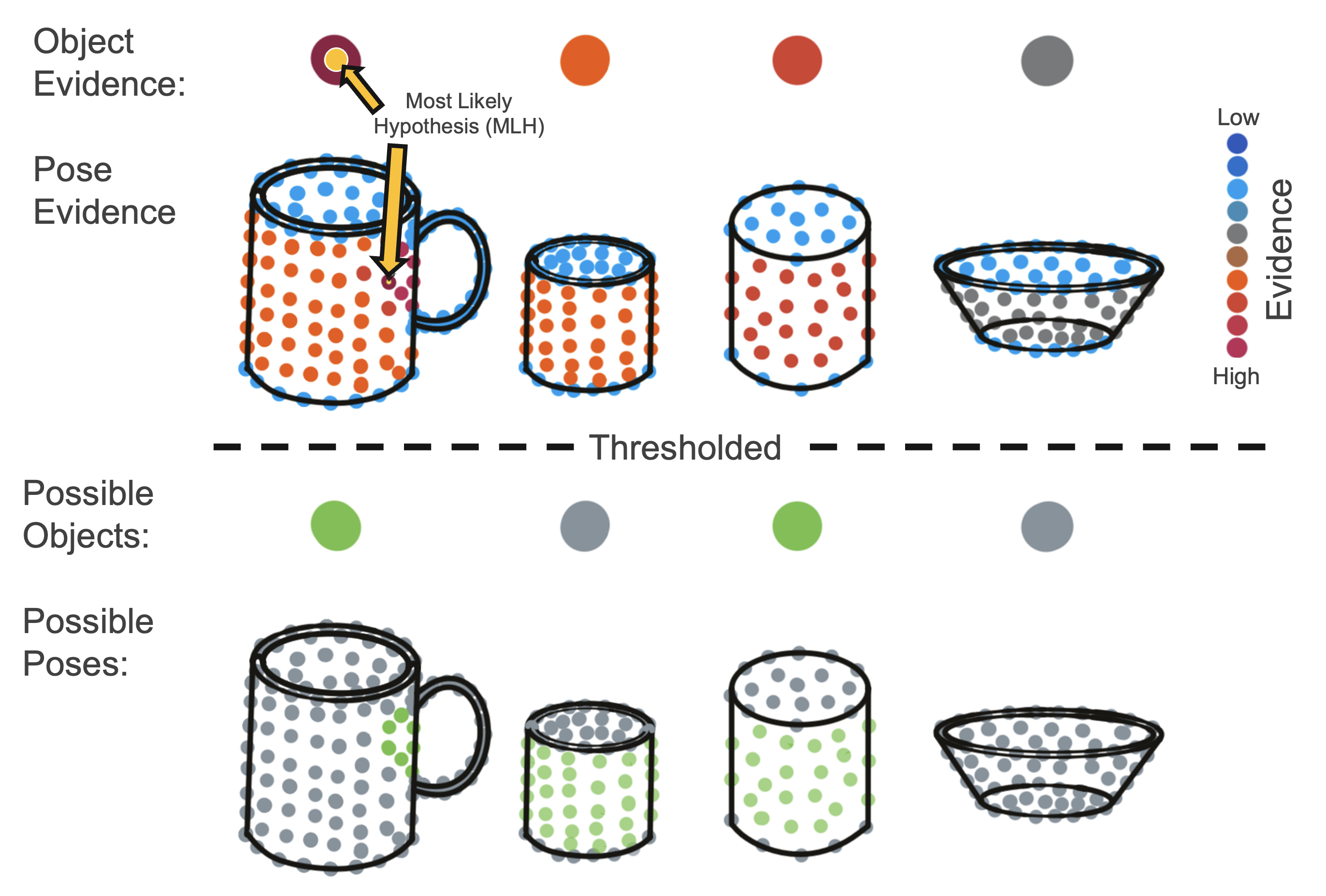}
  \end{center}
  \vspace{-10pt}
  \caption{Thresholding evidence values to obtain possible matches and poses. The highest pose evidence within an object becomes the object's evidence (top row). The highest evidence over all objects is the LM's current most likely hypothesis (MLH, indicated by a yellow dot). We then apply the percent threshold to objects and poses separately (bottom row). In this case, objects 0 and 2 are considered possible. Object 1 has an evidence count that is too low compared to the most likely object. Object 3 has no evidence above 0 and is therefore automatically not possible. We apply the same procedure to the poses. Possible poses for objects 1 and 2 are more transparent as, in practice, these are only calculated for the most likely object.}
  \label{fig:possiblematches}
\end{figure*}

Besides possible matches and possible poses, we also have the most likely hypothesis. This is simply the maximum evidence of all poses, irrespective of any threshold. The most likely hypothesis within one object defines this object's overall evidence and the most likely hypothesis overall (the output of the LM) is the maximum evidence value across all objects and poses. Each LM has a most likely hypothesis at every step, even if it is not confident enough yet to make a classification.

Finally, when an object has no hypothesis with a positive evidence count, it is not considered a possible match. If all objects have only negative evidence, then we do not know the object we are presented with and the LM creates a new model for it in memory.

\subsection{Voting with Evidence}
\begin{figure*}[h]
  \begin{center}
      \includegraphics[width=\textwidth]{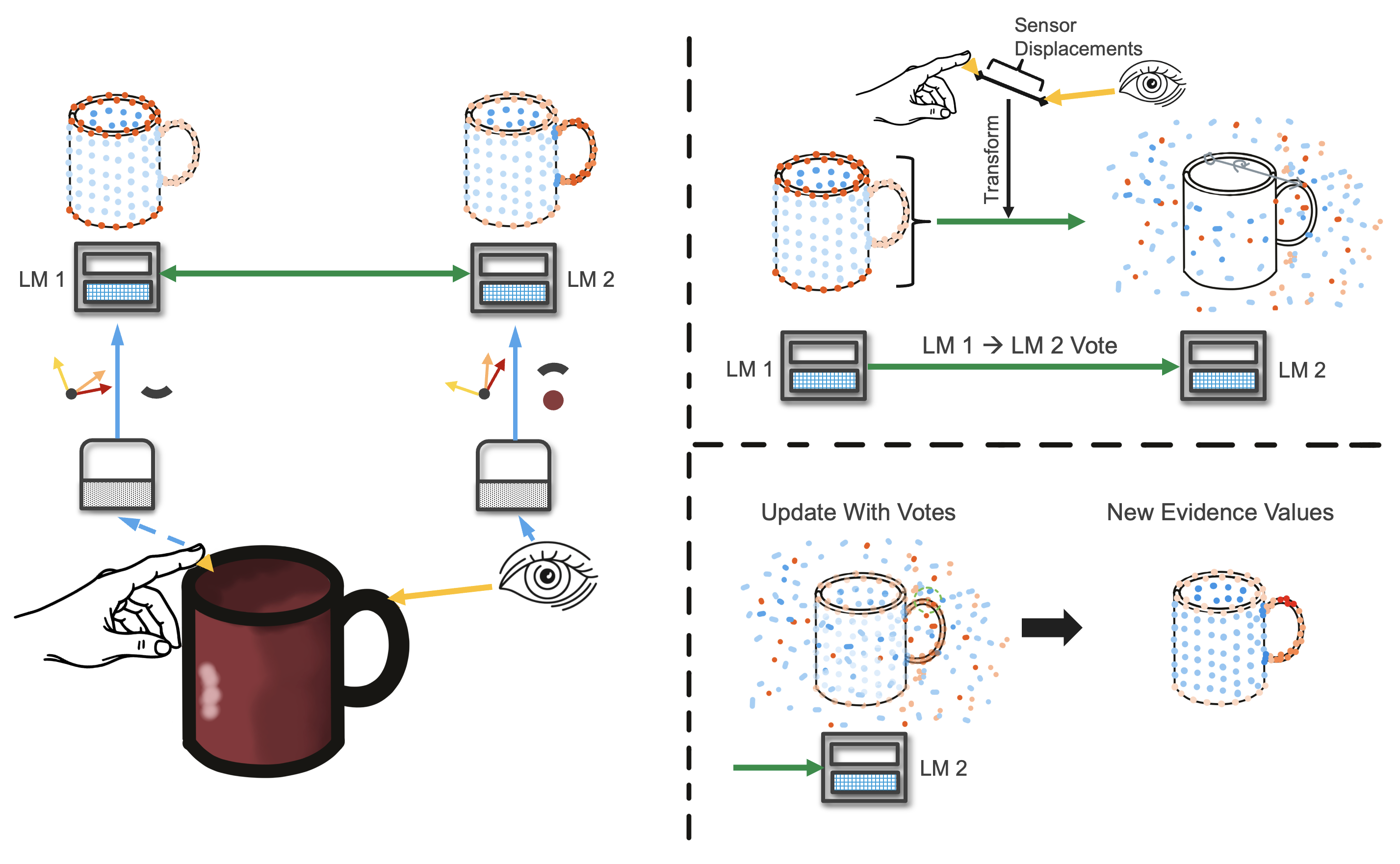}
  \end{center}
  \vspace{-10pt}
  \caption{Vote sent from LM 1 to LM 2 and used to update the evidence in LM 2. The two LMs receive input from two different poses on the mug and two different SMs (left). The possible poses and their evidence in LM 1 are sent to LM 2. In order to put them into the reference frame of LM 2 we need to account for the relative displacement between SM 1 and SM 2 and transform the poses accordingly (top right). This transformation is performed in a common reference-frame (e.g. body-centric). The hypotheses of LM 2 are then updated by looking at the nearest neighbor votes of each (bottom right). After voting we have a clear most likely pose in LM 2 that is consistent with the inputs from SM 1 and SM 2.}
  \label{fig:voting}
\end{figure*}

Voting can help to recognize objects faster as it helps integrate information from multiple matches. In particular, a learning module is able to recognize objects on its own simply through successive movements, however many such movements may be required. With voting we can perform `flash inference' by sharing information between multiple learning modules. Note that, as shown in figure \ref{fig:voting}, voting also works across modalities. This is because votes only contain information about possible objects and their poses which is modality agnostic. At no point does an LM communicate object models or modality specific features to other LMs.

\label{sec:evidencevoting}

At each step, after an LM has updated its evidence based on the current observation, the LM sends out a vote to all its connected LMs. This vote contains its pose hypotheses and the current evidence for each hypothesis. The evidence values are scaled to [-1, 1] where -1 is the currently lowest evidence and 1 is the highest. This makes sure that LMs that received more observations than others do not have an outsized influence. We can also choose to transmit only some votes by sub-selecting those with a proportionally higher evidence value, significantly reducing the computational cost of voting.

The votes get transformed using the displacement between the sensed input poses. We assume that the models in both LMs were learned at the same time and are therefore in the same reference frame. If this does not hold, the reference transform between the models would also have to be applied here, for example, via a learned association.

Once the votes are in the receiving LM's reference frame, the receiving LM updates its evidence values. To do this, it again looks at the nearest neighbor to each hypothesis location, but this time the nearest neighbors in the votes. The distance-weighted average of votes in the search radius (between -1 and 1) is added to the hypothesis evidence.

\subsection{Terminal Conditions}
\label{sec:terminalcond}
In our current experimental setup, we divide time into episodes. Each episode ends when a terminal state is reached. In the object recognition task, this is either \textit{no match} (the model does not know the current object and we construct a new graph for it), \textit{match} (we recognized an object as corresponding to a graph in memory), or \textit{time out} (we took a maximum number of steps without reaching one of the other terminal states). As a result, any given episode contains a variable number of steps, and after every step, we need to check if a terminal condition was met. More details on how the terminal condition is determined are available in our online documentation at \url{https://thousandbrainsproject.readme.io/docs/evidence-based-learning-module#terminal-condition}.

Note that an individual LM can reach its terminal state earlier than the overall Monty system. For example, if one LM does not have a model of the shown object, it will quickly reach the no-match state, while the other LMs will continue until they recognize the object. The episode only ends once a parameter-defined minimum number of LMs have reached their terminal state.

\subsection{Connecting LMs into a Heterarchy}
\subsubsection{Why Heterarchy?}
We use the term \textit{heterarchy} to express the notion that information flow in Monty does not follow a strict hierarchy. In addition to classical hierarchical connections, Monty also has several non-hierarchical forms of connectivity, analogous to long-range connections in the neocortex. Even though we do speak of lower-level LMs and higher-level LMs at times, this does not mean that information flows in a rigid manner from layer 0 to layer N. 

Firstly, there can exit `skip connections' in the network. A low-level LM or even an SM can directly connect to another LM which represents far more complex, high-level models. As such, it is difficult to clearly identify what "layer" an LM belongs to based on the number of previous processing steps performed on its input. Instead, LMs can be grouped into collections based on which LMs vote with one another, which is defined by whether there is (learned) overlap in the objects they model. In other words, voting through lateral connections can also occur between LMs that might classically be viewed as existing at different levels of a hierarchical system.

Second, there exist several channels of communication in Monty that do not implement a hierarchical passing of information (see figure \ref{fig:overview}). An LM can receive multiple top-down signals, which again may bypass certain LMs. These top-down inputs arise from LMs that model compositional objects, and can provide biasing context. In addition to supporting inference, top-down inputs can carry goal states, used for decomposing hierarchical action policies. 

Lastly, each LM can send motor outputs directly to the motor system. This is contrary to the idea that sensory input is processed through a series of hierarchical steps until it reaches a single motor area, which then produces actions. Instead, similar to cortical columns in the brain, each LM in Monty operates as a complete sensorimotor unit. Motor output is not exclusive to the top of a hierarchy but rather occurs at every level of sensory processing.

While the term heterarchy is useful to capture this flexibility in the connectivity of the architecture, it can also be useful to use more traditional terms from hierarchy to aid understanding. Below we provide further details on the bottom-up and top-down information flow in the system.

\subsubsection{Bottom-up connections}
Connections we refer to as bottom-up are connections from SMs to LMs, and connections between LMs that communicate an LMs output (the current most likely object ID and pose) to the main input channel of another LM (the current sensed feature and pose). The output object ID of the sending LM then becomes a feature in the models learned in the receiving LM. For example, the sending LM might be modeling a car tire. When the tire model is recognized, it outputs this and the recognized location and orientation of the tire relative to the body. The receiving LM would not get any information about the 3D structure of the tire from the sending LM. It would only receive the object ID (as a feature) and its pose. This LM could then model a car, composed of different parts. Each part, like the tire, is modeled in detail in a lower-level LM and then becomes a feature in the higher-level LMs' model of the car. 

The receiving LM might additionally get input from other LMs and SMs. For example, the LM modeling the car could also receive direct, low-frequency input from a sensor module and incorporate this into its model. This input, however, is usually not as detailed as the input to the LM that models the tire. In particular, we do not want the higher-level LM to relearn a detailed model of the entire car. Instead, we want to learn detailed models of its components and then compose the components into a larger model. This way, we can also reuse the model of the tire in other higher-level models, such as for trucks, busses, and wheel barrels.

\subsubsection{Top-down connections}
Top-down connections can bias the hypothesis space of the receiving LM, similar to the influence of votes. For example, if a higher-level LM recognizes a car, this can bias the lower-level LMs to recognize the components of a car at a particular location and pose. This enables faster inference and better predictions at lower levels of the system.

\section{Action Policies}
\label{sec:policies}

As noted earlier, policies in thousand-brains systems can be broken down into those that are model-free and those that are model-based. They enable the system to perform rapid, principled actions, and in the future, will enable it to influence the state of the world. Before we discuss these however, it is worth discussing the different \textit{agents} that can currently be leveraged by Monty. These agents in turn have different action spaces, corresponding to their physical properties.

\subsection{Agents and Action Spaces}

We currently have two broad types of agents, along with their particular action spaces. The first is what we call the \textit{distant agent}, which is physically separated from the surface of the object it is sensing. Like an eye, the action space is that of a ball-and-socket actuator that can look in different directions.

The second agent is designed to move freely through space, while being constrained to follow along the surface of an object (like a finger). Hence we call it the \textit{surface agent}. It can efficiently move around the entire object by following its surface, theoretically reaching any feature on the object.

Both the surface and distant agent can be paired with a model-based policy that allows them to make instantaneous "jumps" in the absolute coordinates of the environment. These jumps are currently initiated by model-based policies derived from the LMs, as a form of top-down control.

\begin{figure*}[t]
  \begin{center}
      \includegraphics[width=0.8\textwidth]{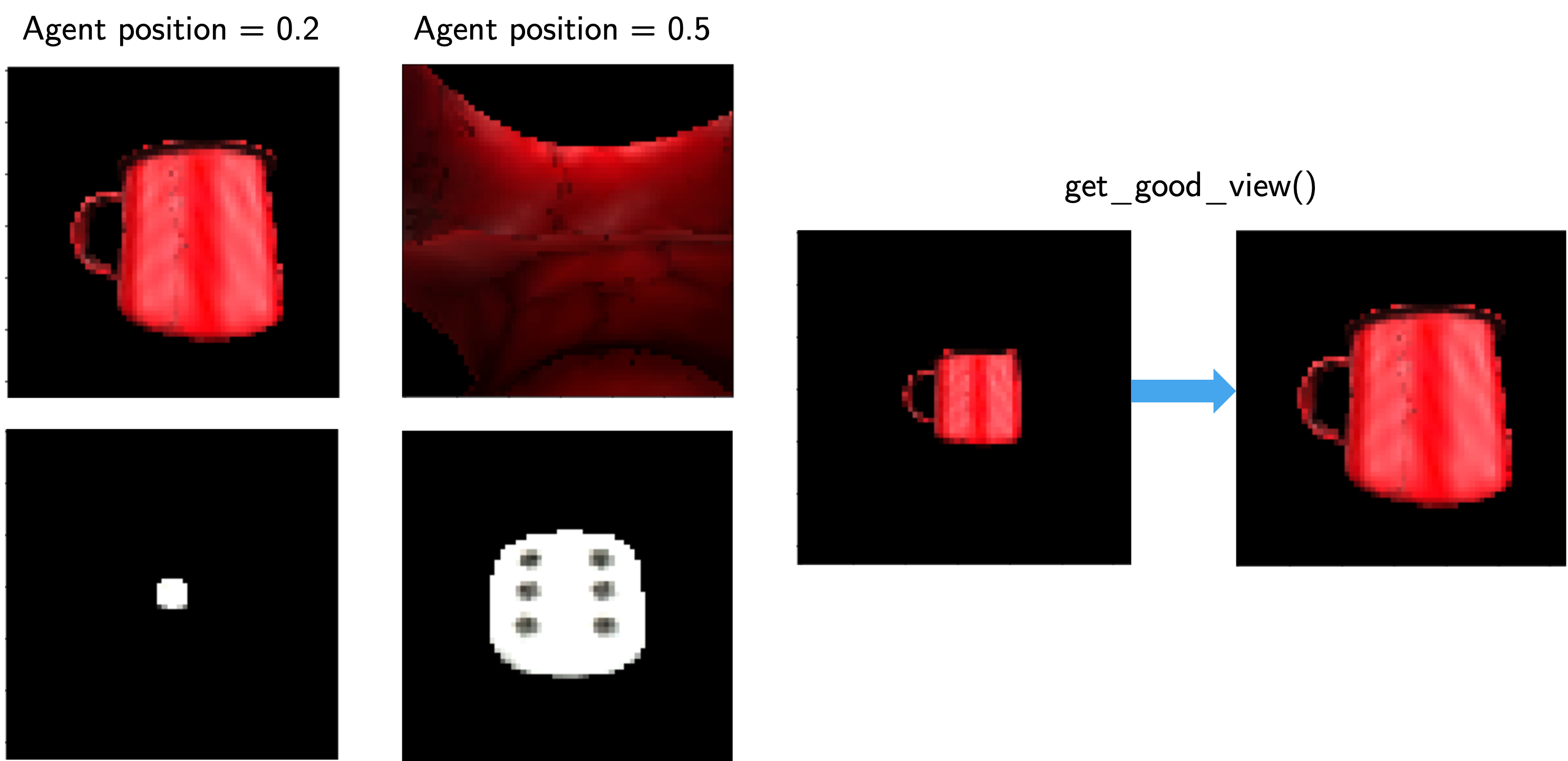}
  \end{center}
  \vspace{-10pt}
  \caption{Using the same object and agent positions for all objects leads to objects covering different amounts of the sensor view (left). The \textit{get\_good\_view} function of the motor system is called once at the beginning of each episode and makes sure that each object covers a similar amount of space in the view-finder (right).}
  \label{fig:goodview}
\end{figure*}

While the distant and surface agent were inspired by an eye vs. a finger sensing the world, in our simulations they are both connected to the same sensor, an RGBD camera. More details on the unique properties that these two action spaces afford are covered in our online documentation at \url{https://thousandbrainsproject.readme.io/docs/policy}.

\subsection{Utility Policies}

Before an experiment starts, the agent is moved to an appropriate starting position relative to the object. This serves to set up the conditions desired by the human operator and is analogous to a neurophysiologist lifting an animal to place it in a particular location and orientation in a lab environment. As such, these are considered utility functions or "policies", in that they are not driven by the intelligence of Monty, although they currently make use of its internal action spaces. Furthermore, sensory observations that occur during the execution of a utility policy are not sent to the learning module(s), as they have access to privileged information, such as a wider field-of-view camera. Two such policies exist, one for the distant agent ("get good view"), and one for the surface agent ("touch object").

For the former, the distant agent is moved to a "good view" such that small and large objects in the data set cover approximately a similar space in the camera image (see Figure \ref{fig:goodview}). To determine a good view, we use the view-finder, which is a camera without zoom that sees a larger picture than the sensor patch. Without this policy, small objects, such as the dice, may be smaller than the sensor patch itself, thereby preventing any movement of the sensor patch on the object. For large objects, there is a risk that the agent is initialized inside the object, as shown in the second image in the first row of the figure below. 

The "touch object" policy serves a similar purpose - the surface agent is moved sufficiently close such that it is essentially on the surface of the object. This will be important in future work when the surface agent has access to sensory inputs, such as texture, that require maintaining physical contact with an object.

\subsection{Model-Free and Model-Based Policies}

Turning to the core policies of Monty, we can divide these into two broad categories: model-based policies where actions are determined using the internal, structured models of LMs, and those that only rely on sensory inputs and are therefore model-free. A special case of the model-free policy is the random policy that is not influenced by any factors. While simpler, model-free policies can already provide significant benefits to the efficiency of a system, while introducing minimal computational overhead. On the other hand, model-based policies make use of the learned object models and current hypotheses to support deliberate, intelligent actions. Importantly, model-free and model-based policies can work in concert for maximal efficiency.

\begin{figure}[t]
  \begin{center}
      \includegraphics[width=0.4\textwidth]{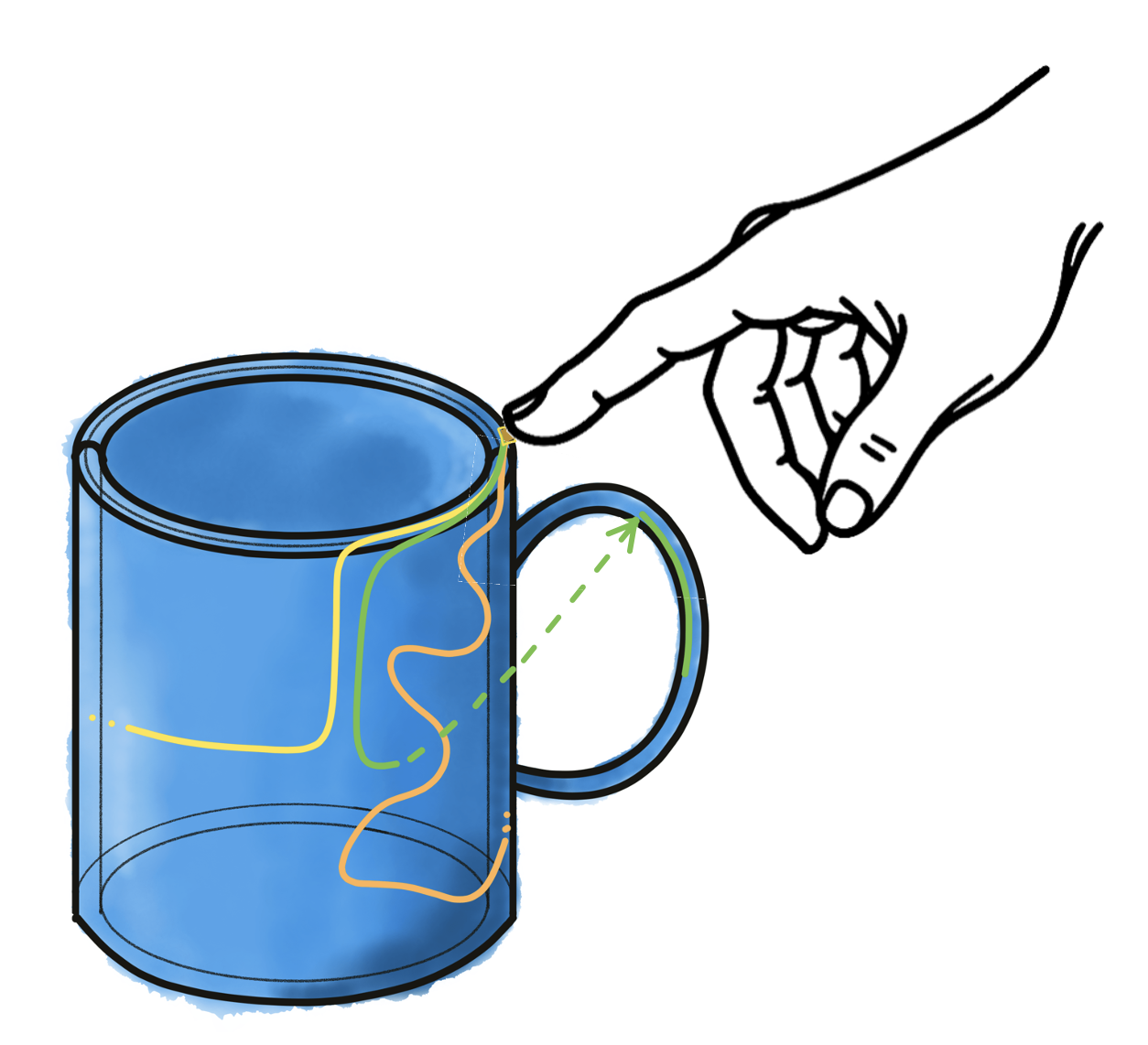}
  \end{center}
  \vspace{-10pt}
  \caption{Comparison of different surface-agent policies. (orange) Random movement along the object's surface. (yellow) model-free policy that follows principal curvature directions. (green) model-based policy that can jump to specific locations on the object to actively test and contrast the current most likely hypotheses.}
  \label{fig:policies}
\end{figure}

Model-free policies make use of sensory input from the SM. These can be compared to the actions that humans perform primarily through sub-cortical structures, such as the motor control required for walking, or balancing while riding a bicycle. They can be innate, such as reflexes related to potentially dangerous stimuli, or learned.

Model-based policies are based on \textit{goal-states} that are output from the goal-state generator of the learning modules. These goal states follow the CMP and therefore contain a pose and features. This information is interpreted as a target pose of an object that should be achieved in the world and is translated into motor commands in the motor system.

\subsubsection{Concrete Model-based and Model-Free Policies Used in Monty}
In the current distant agent, observations are collected by random movement of the camera. The only model-free influence is that if the sensor patch moves off the object, the previous action is reversed to make sure we stay on the object. Policies with random elements also have a momentum parameter (alpha) that regulates how likely it is to repeat the previous action and allows for more directional movement paths.

The surface agent can either use a random walk policy (again with an optional momentum parameter to bias following a consistent path), or alternatively make use of the "curvature-informed" policy. This policy makes use of sensed principle curvature directions, attempting to follow these where they are present, such as on the rim of a cup, or the handle of a mug. The details of this policy are expanded upon further below. 

Finally, both the distant and surface agent can be controlled by a model-based policy, what is called the hypothesis-testing policy.

\subsection{Policies for Inference vs. Learning. vs. Manipulating the Environment}

The policies mentioned above are aimed at efficient inference, and constitute the focus of our evaluation environments to date. There is also a specialized policy that can be used to ensure sufficient object coverage when the distant agent is learning about new objects, called the scan policy. This policy starts at the center of the object and moves outwards on a spiral path. In addition, model-based policies can also be leveraged to make learning more efficient, such as directing sensory systems to areas of the world that are under-represented in internal models. As such, this will be a subject of future research. Finally, the long-term aim is that thousand-brains systems not only quickly understand their environment, but can also mediate change in the state of the world. As such, action policies that enable the system to interact with external objects will be a major focus of future work. 

\begin{figure*}[t]
  \begin{center}
      \includegraphics[width=0.8\textwidth]{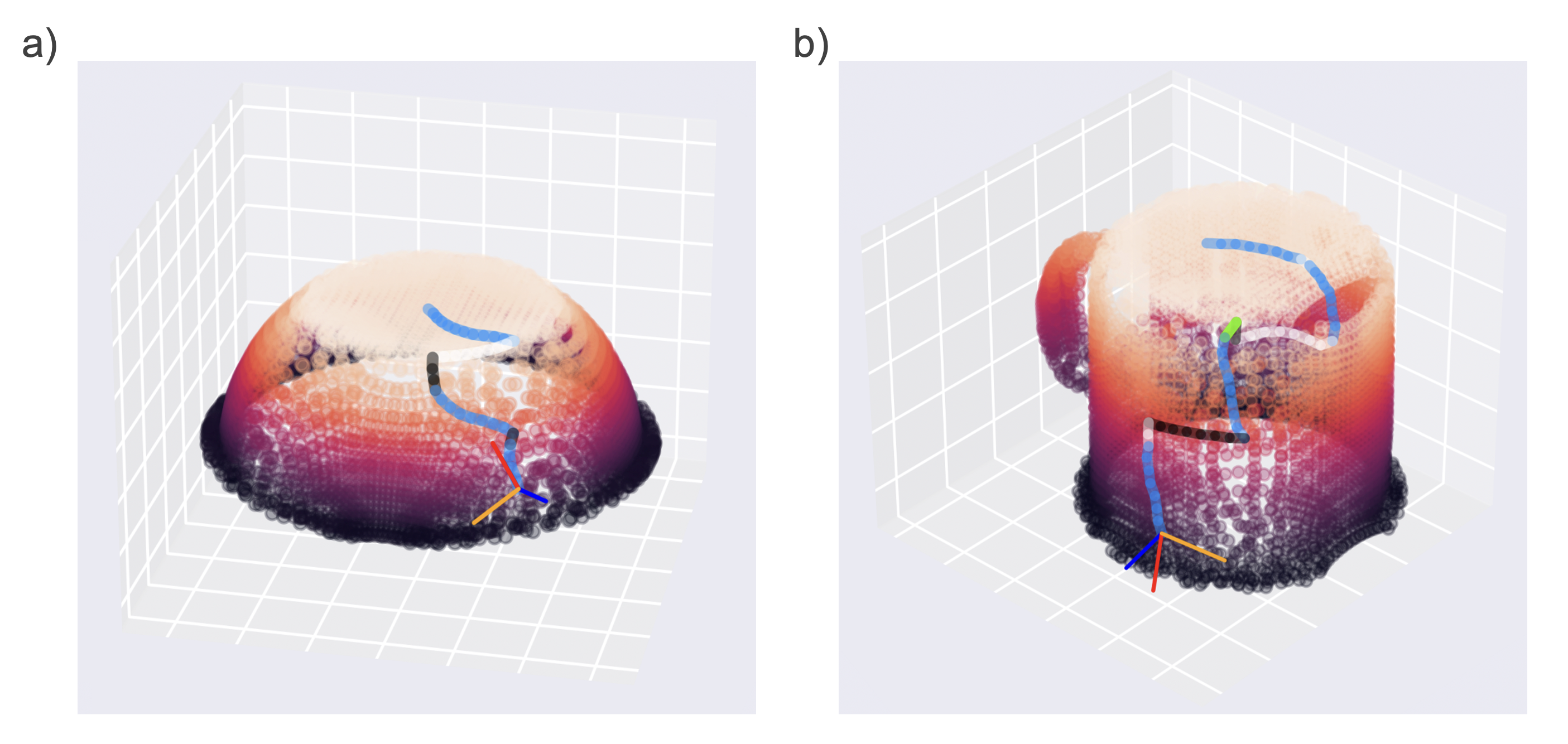}
  \end{center}
  \vspace{-10pt}
  \caption{Two samples of the curvature-informed surface policy being used during inference. Blue segments represent moving randomly with some momentum, white represents following minimal principal curvature, black following maximal, and green taking an avoidance step. Note in particular how in (b), as the agent moves over the rim of the cup, it realizes it will revisit a previous location, and so takes an avoidance step (green) to bring it in a new direction. Further note that even when principal curvature might be evident to the human eye, the model may not be receiving a valid PC-input due to noise on the surface of the mesh. Finally, the end point shows the sensed orientation of the current feature (blue detected point-normal, red and orange detected principal curvatures).}
  \label{fig:curve_pol_example}
\end{figure*}

\begin{figure*}[t]
  \begin{center}
      \includegraphics[width=1.0\textwidth]{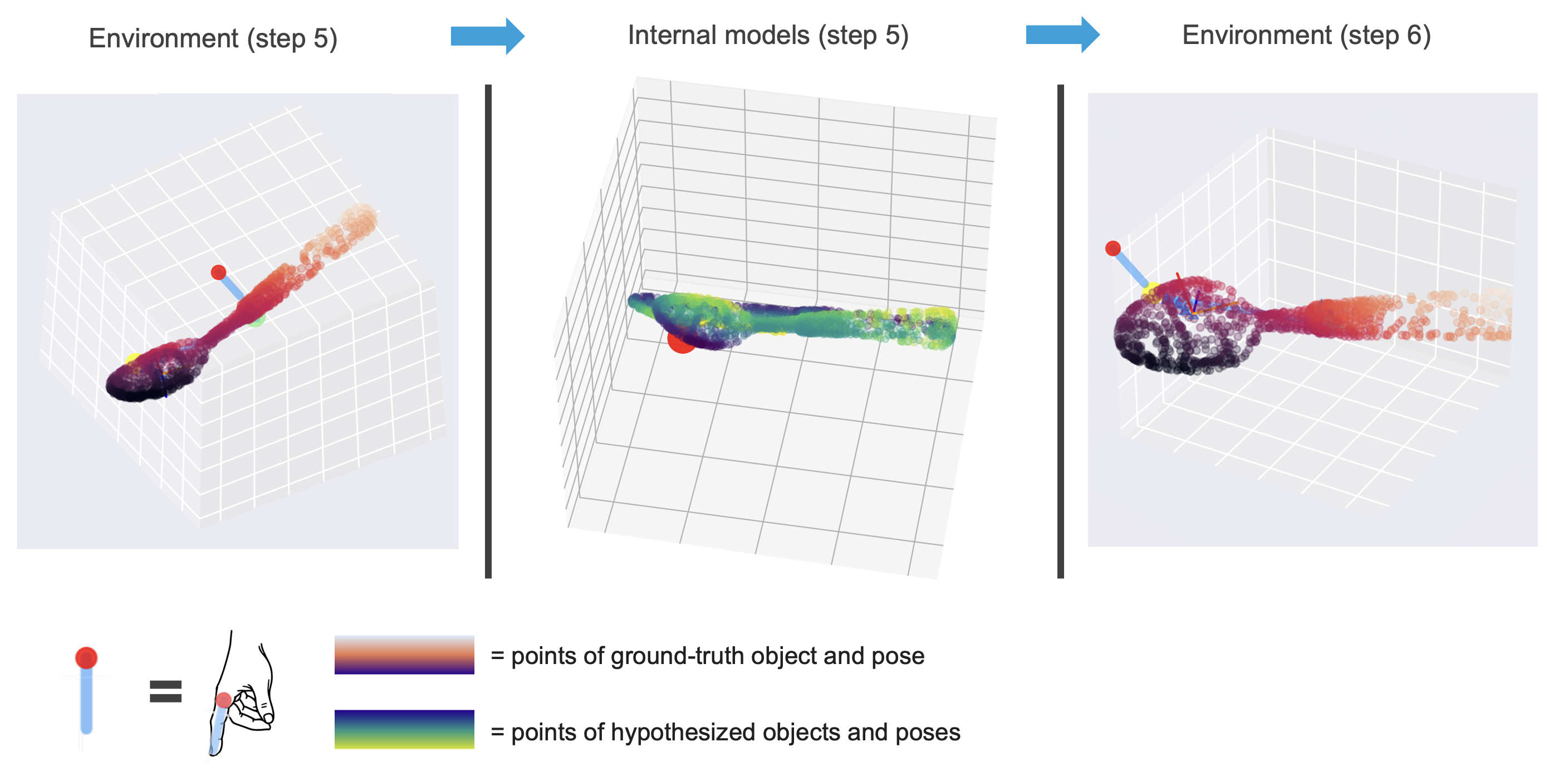}
  \end{center}
  \vspace{-10pt}
  \caption{The hypothesis-testing policy. The left and right plots show where the agent is in the external world at two different time-steps (surface agent represented with ball-and-pole, where the pole points in the direction the agent is facing). The central plot shows the LMs internal model of the hypothesis spaces. The LM uses its estimate of its current location and pose on the two most-likely objects (spoon and knife) to compare their relative orientations in its internal "mental" space. Once overlaid, the graph-mismatch technique proposes testing a part of the head of the spoon (red-spot, center) as it maximally distinguishes that graph from the other. Here Euclidean distance in 3D space is used, but this process can also be performed in feature space. In the final panel, we see the agent after instantaneously moving to the test point.}
  \label{fig:hyp_driven_example}
\end{figure*}

\subsubsection{Policy Algorithm Details}

Due to their relative complexity, the below sub-sections provide further detail on the curvature-informed and hypothesis-testing policies.

\textbf{Curvature Informed Policy (Model-Free)}

The curvature-informed surface-agent policy is designed to follow the principal curvatures on an object's surface (where these are present), so as to efficiently explore the relevant "parts" of an object. For example, this policy should be more likely than a random walk with momentum to efficiently explore components such as the rim of a cup, or the handle of a mug, when these are encountered. 

To enable this, the policy is informed by the principal curvature information available from the sensor-module. It will alternate between following the minimal curvature (e.g. the rim of a cup) and the maximal curvature (e.g. the curved side of a cylinder). The decision process that guides the curvature-guided policy is elaborated in our online documentation (\url{https://thousandbrainsproject.readme.io/docs/policy#curvature-informed-policy-details}). We also go into detail there about the ability of the curvature-informed policy to avoid previously sampled locations, a capability that will likely be implemented as a model-based policy in the future.

In Figure \ref{fig:curve_pol_example}, two examples of the policy in action are shown. 

\textbf{Hypothesis-Testing Policy (Model-Based)}

For the hypothesis-testing action policy, Monty uses its learned internal models of objects. In particular, an LM's learned models enable hypotheses about the current ID and pose of the object that it is perceiving. By comparing the models of the most likely hypotheses, the hypothesis-testing policy enables an LM to propose a point in space to move to that can rapidly disambiguate the actual object observed. 

To determine the most distinguishing part of the object to test, a form of "graph mismatch" is employed. This technique takes the most likely and second most likely object graphs, and using their most likely poses, overlays them in an internal ("mental") space. It then determines for every point in the most likely graph, how far the nearest neighbor is in the second graph. The output is the point in the first graph that has the most distant nearest neighbor (Figure \ref{fig:hyp_driven_example}).

As a model-based policy, the LM communicates the desired action in the form of a goal-state. This goal-state represents a state that the motor-system should achieve, such as sensing the point of interest with a sensor-augmented actuator.

Bringing this together, consider the example of recognizing a mug, where the LM also knows about other cylindrical objects such as cans. If the most likely object was a mug, and the second most likely object a can of soup, then a point on the handle of the mug would have the most-distant nearest neighbor to the can-of-soup graph. The LM would communicate to the motor systems a goal-state to be observing the point in body-centric coordinates corresponding to the handle of the mug. In this way, the LM can coordinate intelligent behavior in the motor-system, without the motor-system needing to know anything about objects like mugs or soup cans.

For the graph-mismatch component, using the Euclidean distance between graph points is a reasonable heuristic for identifying potentially diagnostic differences in the structures of two objects. However, in the future we may also want to look at distances in feature space, which will be particularly interesting once features represent sub-components in higher-level LMs.

In addition to being able to compare the top two most likely objects in such a way, one can just focus on the most likely object, and use this same technique to compare the two most likely \textit{poses} of the same object. 

Details for the decision-process that determines exactly when a hypothesis-testing action is generated by an LM are provided in our online documentation at \url{https://thousandbrainsproject.readme.io/docs/policy#hypothesis-driven-policy-details}.

\subsection{Long Term Policy View}

\subsubsection{Learning, and the Interplay of Model-Free and Model-Based Policies}

It is worth emphasizing that future versions of Monty will certainly leverage various advanced learning techniques, including reinforcement learning. It is important to note therefore that our intent is not to explicitly implement policies for every conceivable scenario an agent might face. Rather, the policies we are developing are intended as a reasonable set of primitives that other policies might make use of, without having to learn them from the ground up. This is analogous to examples such as i) the tendency of infants to attend to human faces (innate, model-free, ref) ii) the structural knowledge of bicycles and people that enables a child to observe an adult bicycling, and thereby efficiently attempt peddling (learned, model-based) iii) the fine-motor coordination and balance that enables proficiently riding a bicycle (learned, model-free).

\subsubsection{Abstract Spaces}

It is also worth noting that the policy primitives we are developing will likely prove useful as we move from 3D physical space to more abstract spaces. For example, in such settings, we might still want to have an inductive bias that moves along dimensions of maximal or minimal variation of a manifold, akin to the curvature-guided surface policy. Similarly, the mental alignment of structured representations to identify the most relevant points that distinguish different concepts would be useful in a variety of abstract settings, such as comparing computational algorithms.

\subsection{Hierarchical Model-Based Policies}

Finally, the model-based policy we have so far described makes use of only a single level of reference-frame based representations. We are in the process of implementing hierarchical, compositional representations, which will also enable hierarchical, model-based policies. In particular, the goal-states that an LM outputs can also be received by other LMs. This affords the ability to decompose a complex task into simpler tasks, where an LM is responsible for a given task as a function of what it knows about the world. For example, at the highest level of a system there might be an innate requirement, such as a person wishing to improve their energy levels. An LM that models the way alertness can be increased would have a goal of making a cup of coffee. This goal-state could be sent to an LM that models a kitchen environment, which will in turn recruit an LM that models the structure and behavior of coffee machines. Eventually, such goal-states will always decompose into a simple goal-state that can be sent directly to motor-systems. The signal to the motor-system is analogous to the direct, motor projections found in all cortical columns of the cortex, while hierarchical goal-states likely correspond to the connectivity between cortical columns.

Some novel actions may require complex, model-based planning of a system's own actuators, such as trying to push a computer mouse with your elbow, or holding a pen between the knuckles of your left and right ring-fingers. This will likely require LMs that model actuator objects (e.g. robotic limbs), rather than objects external to the system, analogous to the motor cortex of the brain. Such systems might receive goal-states from other regions in the system, before ultimately recruiting simpler, model-free policies present at the level of the motor system. After many repetitions, policies that were originally model-based and relatively laborious can become model free.

\section{Conclusion}

We have outlined our vision for sensorimotor AI based on the operating principles of the neocortex. Thousand-brains systems replicate a core computational-unit - the cortical column implemented as the learning-module - to model objects in the world in any sensory modality, and at any level of abstraction. Each unit operates as a semi-independent sensorimotor system, but is also able to communicate key information with other learning modules via a common communication protocol. By leveraging structured, internal models, learning-modules can rapidly learn about the world, and leverage this knowledge for sophisticated policies. Building on these core concepts, we described Monty, the first instantiation of a thousand-brains system. While a simple, first-generation implementation, we believe that these same principles will apply to all future thousand-brains systems, affording increasing intelligence as the sophistication of their implementations grows. More broadly, we hold that the principles described here will be the foundation for a new type of AI, one that  efficiently learns generalizable representations from sensorimotor data.

\section{Acknowledgments}

We would like to thank the following individuals for invaluable discussions on the Thousand Brains theory and Monty concepts: Subutai Ahmad, Heiko Hoffmann, and Kevin Hunter. In addition to such contributions to discussions, we would like to thank the following individuals for their contributions to the Monty code-base: Ben Cohen, Jad Hanna, Abhiram Iyer, Ramy Mounir, Luiz Scheinkman, Philip Shamash, and Lucas Souza.

\bibliographystyle{unsrtnat}
\bibliography{references}

\begin{thebibliography}{17}
\providecommand{\natexlab}[1]{#1}
\providecommand{\url}[1]{\texttt{#1}}
\expandafter\ifx\csname urlstyle\endcsname\relax
  \providecommand{\doi}[1]{doi: #1}\else
  \providecommand{\doi}{doi: \begingroup \urlstyle{rm}\Url}\fi

\bibitem[Hawkins et~al.(2019)Hawkins, Lewis, Klukas, Purdy, and Ahmad]{Hawkins2019ANeocortex}
Jeff Hawkins, Marcus Lewis, Mirko Klukas, Scott Purdy, and Subutai Ahmad.
\newblock {A framework for intelligence and cortical function based on grid cells in the neocortex}.
\newblock \emph{Frontiers in Neural Circuits}, 2019.
\newblock ISSN 16625110.
\newblock \doi{10.3389/fncir.2018.00121}.

\bibitem[Mountcastle(1997)]{Mountcastle1997TheNeocortex}
Vernon~B. Mountcastle.
\newblock {The columnar organization of the neocortex}.
\newblock \emph{Brain}, 120\penalty0 (4), 1997.
\newblock ISSN 00068950.
\newblock \doi{10.1093/brain/120.4.701}.

\bibitem[Edelman and Mountcastle(1978)]{Edelman1978MindfulBrain}
G~Edelman and V~Mountcastle.
\newblock The mindful brain: Cortical organization and the group-selective theory of higher brain function.
\newblock \emph{pp}, 100, 1978.
\newblock URL \url{https://psycnet.apa.org/record/1979-25355-000}.

\bibitem[Driess et~al.(2023)Driess, Xia, Sajjadi, Lynch, Chowdhery, Ichter, Wahid, Tompson, Vuong, Yu, Huang, Chebotar, Sermanet, Duckworth, Levine, Vanhoucke, Hausman, Toussaint, Greff, Zeng, Mordatch, and Florence]{Driess2023}
Danny Driess, Fei Xia, Mehdi~S.M. Sajjadi, Corey Lynch, Aakanksha Chowdhery, Brian Ichter, Ayzaan Wahid, Jonathan Tompson, Quan Vuong, Tianhe Yu, Wenlong Huang, Yevgen Chebotar, Pierre Sermanet, Daniel Duckworth, Sergey Levine, Vincent Vanhoucke, Karol Hausman, Marc Toussaint, Klaus Greff, Andy Zeng, Igor Mordatch, and Pete Florence.
\newblock Palm-e: An embodied multimodal language model.
\newblock In \emph{Proceedings of Machine Learning Research}, volume 202, 2023.

\bibitem[{OpenAI} et~al.(2023){OpenAI}, Achiam, Adler, Agarwal, Ahmad, Akkaya, Aleman, Almeida, Altenschmidt, Altman, Anadkat, Avila, Babuschkin, Balaji, Balcom, Baltescu, Bao, Bavarian, Belgum, Bello, Berdine, Bernadett-Shapiro, Berner, Bogdonoff, Boiko, Boyd, Brakman, Brockman, Brooks, Brundage, Button, Cai, Campbell, Cann, Carey, Carlson, Carmichael, Chan, Chang, Chantzis, Chen, Chen, Chen, Chen, Chen, Chess, Cho, Chu, Chung, Cummings, Currier, Dai, Decareaux, Degry, Deutsch, Deville, Dhar, Dohan, Dowling, Dunning, Ecoffet, Eleti, Eloundou, Farhi, Fedus, Felix, Fishman, Forte, Fulford, Gao, Georges, Gibson, Goel, Gogineni, Goh, Gontijo-Lopes, Gordon, Grafstein, Gray, Greene, Gross, Gu, Guo, Hallacy, Han, Harris, He, Heaton, Heidecke, Hesse, Hickey, Hickey, Hoeschele, Houghton, Hsu, Hu, Hu, Huizinga, Jain, Jain, Jang, Jiang, Jiang, Jin, Jin, Jomoto, Jonn, Jun, Kaftan, Kaiser, Kamali, Kanitscheider, Keskar, Khan, Kilpatrick, Kim, Kim, Kim, Kirchner, Kiros, Knight, Kokotajlo, Kondraciuk, Kondrich,
  Konstantinidis, Kosic, Krueger, Kuo, Lampe, Lan, Lee, Leike, Leung, Levy, Li, Lim, Lin, Lin, Litwin, Lopez, Lowe, Lue, Makanju, Malfacini, Manning, Markov, Markovski, Martin, Mayer, Mayne, McGrew, McKinney, McLeavey, McMillan, McNeil, Medina, Mehta, Menick, Metz, Mishchenko, Mishkin, Monaco, Morikawa, Mossing, Mu, Murati, Murk, Mély, Nair, Nakano, Nayak, Neelakantan, Ngo, Noh, Ouyang, O'Keefe, Pachocki, Paino, Palermo, Pantuliano, Parascandolo, Parish, Parparita, Passos, Pavlov, Peng, Perelman, Peres, Petrov, Pinto, {Michael}, {Pokorny}, Pokrass, Pong, Powell, Power, Power, Proehl, Puri, Radford, Rae, Ramesh, Raymond, Real, Rimbach, Ross, Rotsted, Roussez, Ryder, Saltarelli, Sanders, Santurkar, Sastry, Schmidt, Schnurr, Schulman, Selsam, Sheppard, Sherbakov, Shieh, Shoker, Shyam, Sidor, Sigler, Simens, Sitkin, Slama, Sohl, Sokolowsky, Song, Staudacher, Such, Summers, Sutskever, Tang, Tezak, Thompson, Tillet, Tootoonchian, Tseng, Tuggle, Turley, Tworek, Uribe, Vallone, Vijayvergiya, Voss, Wainwright, Wang,
  Wang, Wang, Ward, Wei, Weinmann, Welihinda, Welinder, Weng, Weng, Wiethoff, Willner, Winter, Wolrich, Wong, Workman, Wu, Wu, Wu, Xiao, Xu, Yoo, Yu, Yuan, Zaremba, Zellers, Zhang, Zhang, Zhao, Zheng, Zhuang, Zhuk, and Zoph]{OpenAI2023gpt4}
{OpenAI}, Josh Achiam, Steven Adler, Sandhini Agarwal, Lama Ahmad, Ilge Akkaya, Florencia~Leoni Aleman, Diogo Almeida, Janko Altenschmidt, Sam Altman, Shyamal Anadkat, Red Avila, Igor Babuschkin, Suchir Balaji, Valerie Balcom, Paul Baltescu, Haiming Bao, Mohammad Bavarian, Jeff Belgum, Irwan Bello, Jake Berdine, Gabriel Bernadett-Shapiro, Christopher Berner, Lenny Bogdonoff, Oleg Boiko, Madelaine Boyd, Anna-Luisa Brakman, Greg Brockman, Tim Brooks, Miles Brundage, Kevin Button, Trevor Cai, Rosie Campbell, Andrew Cann, Brittany Carey, Chelsea Carlson, Rory Carmichael, Brooke Chan, Che Chang, Fotis Chantzis, Derek Chen, Sully Chen, Ruby Chen, Jason Chen, Mark Chen, Ben Chess, Chester Cho, Casey Chu, Hyung~Won Chung, Dave Cummings, Jeremiah Currier, Yunxing Dai, Cory Decareaux, Thomas Degry, Noah Deutsch, Damien Deville, Arka Dhar, David Dohan, Steve Dowling, Sheila Dunning, Adrien Ecoffet, Atty Eleti, Tyna Eloundou, David Farhi, Liam Fedus, Niko Felix, Simón~Posada Fishman, Juston Forte, Isabella Fulford, Leo
  Gao, Elie Georges, Christian Gibson, Vik Goel, Tarun Gogineni, Gabriel Goh, Rapha Gontijo-Lopes, Jonathan Gordon, Morgan Grafstein, Scott Gray, Ryan Greene, Joshua Gross, Shixiang~Shane Gu, Yufei Guo, Chris Hallacy, Jesse Han, Jeff Harris, Yuchen He, Mike Heaton, Johannes Heidecke, Chris Hesse, Alan Hickey, Wade Hickey, Peter Hoeschele, Brandon Houghton, Kenny Hsu, Shengli Hu, Xin Hu, Joost Huizinga, Shantanu Jain, Shawn Jain, Joanne Jang, Angela Jiang, Roger Jiang, Haozhun Jin, Denny Jin, Shino Jomoto, Billie Jonn, Heewoo Jun, Tomer Kaftan, Łukasz Kaiser, Ali Kamali, Ingmar Kanitscheider, Nitish~Shirish Keskar, Tabarak Khan, Logan Kilpatrick, Jong~Wook Kim, Christina Kim, Yongjik Kim, Jan~Hendrik Kirchner, Jamie Kiros, Matt Knight, Daniel Kokotajlo, Łukasz Kondraciuk, Andrew Kondrich, Aris Konstantinidis, Kyle Kosic, Gretchen Krueger, Vishal Kuo, Michael Lampe, Ikai Lan, Teddy Lee, Jan Leike, Jade Leung, Daniel Levy, Chak~Ming Li, Rachel Lim, Molly Lin, Stephanie Lin, Mateusz Litwin, Theresa Lopez, Ryan
  Lowe, Patricia Lue, Anna Makanju, Kim Malfacini, Sam Manning, Todor Markov, Yaniv Markovski, Bianca Martin, Katie Mayer, Andrew Mayne, Bob McGrew, Scott~Mayer McKinney, Christine McLeavey, Paul McMillan, Jake McNeil, David Medina, Aalok Mehta, Jacob Menick, Luke Metz, Andrey Mishchenko, Pamela Mishkin, Vinnie Monaco, Evan Morikawa, Daniel Mossing, Tong Mu, Mira Murati, Oleg Murk, David Mély, Ashvin Nair, Reiichiro Nakano, Rajeev Nayak, Arvind Neelakantan, Richard Ngo, Hyeonwoo Noh, Long Ouyang, Cullen O'Keefe, Jakub Pachocki, Alex Paino, Joe Palermo, Ashley Pantuliano, Giambattista Parascandolo, Joel Parish, Emy Parparita, Alex Passos, Mikhail Pavlov, Andrew Peng, Adam Perelman, Filipe de Avila~Belbute Peres, Michael Petrov, Henrique Ponde de~Oliveira Pinto, {Michael}, {Pokorny}, Michelle Pokrass, Vitchyr~H Pong, Tolly Powell, Alethea Power, Boris Power, Elizabeth Proehl, Raul Puri, Alec Radford, Jack Rae, Aditya Ramesh, Cameron Raymond, Francis Real, Kendra Rimbach, Carl Ross, Bob Rotsted, Henri Roussez,
  Nick Ryder, Mario Saltarelli, Ted Sanders, Shibani Santurkar, Girish Sastry, Heather Schmidt, David Schnurr, John Schulman, Daniel Selsam, Kyla Sheppard, Toki Sherbakov, Jessica Shieh, Sarah Shoker, Pranav Shyam, Szymon Sidor, Eric Sigler, Maddie Simens, Jordan Sitkin, Katarina Slama, Ian Sohl, Benjamin Sokolowsky, Yang Song, Natalie Staudacher, Felipe~Petroski Such, Natalie Summers, Ilya Sutskever, Jie Tang, Nikolas Tezak, Madeleine~B Thompson, Phil Tillet, Amin Tootoonchian, Elizabeth Tseng, Preston Tuggle, Nick Turley, Jerry Tworek, Juan Felipe~Cerón Uribe, Andrea Vallone, Arun Vijayvergiya, Chelsea Voss, Carroll Wainwright, Justin~Jay Wang, Alvin Wang, Ben Wang, Jonathan Ward, Jason Wei, C~J Weinmann, Akila Welihinda, Peter Welinder, Jiayi Weng, Lilian Weng, Matt Wiethoff, Dave Willner, Clemens Winter, Samuel Wolrich, Hannah Wong, Lauren Workman, Sherwin Wu, Jeff Wu, Michael Wu, Kai Xiao, Tao Xu, Sarah Yoo, Kevin Yu, Qiming Yuan, Wojciech Zaremba, Rowan Zellers, Chong Zhang, Marvin Zhang, Shengjia
  Zhao, Tianhao Zheng, Juntang Zhuang, William Zhuk, and Barret Zoph.
\newblock {GPT}-4 technical report.
\newblock \emph{arXiv [cs.CL]}, March 2023.
\newblock URL \url{http://arxiv.org/abs/2303.08774}.

\bibitem[Black et~al.(2024)Black, Brown, Driess, Esmail, Equi, Finn, Fusai, Groom, Hausman, Ichter, Jakubczak, Jones, Ke, Levine, Li-Bell, Mothukuri, Nair, Pertsch, Shi, Tanner, Vuong, Walling, Wang, and Zhilinsky]{Black2024}
Kevin Black, Noah Brown, Danny Driess, Adnan Esmail, Michael Equi, Chelsea Finn, Niccolo Fusai, Lachy Groom, Karol Hausman, Brian Ichter, Szymon Jakubczak, Tim Jones, Liyiming Ke, Sergey Levine, Adrian Li-Bell, Mohith Mothukuri, Suraj Nair, Karl Pertsch, Lucy~Xiaoyang Shi, James Tanner, Quan Vuong, Anna Walling, Haohuan Wang, and Ury Zhilinsky.
\newblock pi0: A vision-language-action flow model for general robot control.
\newblock \emph{arXiv}, 10 2024.
\newblock URL \url{http://arxiv.org/abs/2410.24164}.

\bibitem[Reed et~al.(2022)Reed, Zolna, Parisotto, Colmenarejo, Novikov, Barth-Maron, Gimenez, Sulsky, Kay, Springenberg, Eccles, Bruce, Razavi, Edwards, Heess, Chen, Hadsell, Vinyals, Bordbar, and de~Freitas]{Reed2022}
Scott Reed, Konrad Zolna, Emilio Parisotto, Sergio~Gomez Colmenarejo, Alexander Novikov, Gabriel Barth-Maron, Mai Gimenez, Yury Sulsky, Jackie Kay, Jost~Tobias Springenberg, Tom Eccles, Jake Bruce, Ali Razavi, Ashley Edwards, Nicolas Heess, Yutian Chen, Raia Hadsell, Oriol Vinyals, Mahyar Bordbar, and Nando de~Freitas.
\newblock A generalist agent.
\newblock \emph{Transactions on Machine Learning Research}, 5 2022.
\newblock URL \url{http://arxiv.org/abs/2205.06175}.

\bibitem[Team et~al.(2024)Team, Raad, Ahuja, Barros, Besse, Bolt, Bolton, Brownfield, Buttimore, Cant, Chakera, Chan, Clune, Collister, Copeman, Cullum, Dasgupta, de~Cesare, Trapani, Donchev, Dunleavy, Engelcke, Faulkner, Garcia, Gbadamosi, Gong, Gonzales, Gupta, Gregor, Hallingstad, Harley, Haves, Hill, Hirst, Hudson, Hudson, Hughes-Fitt, Rezende, Jasarevic, Kampis, Ke, Keck, Kim, Knagg, Kopparapu, Lawton, Lampinen, Legg, Lerchner, Limont, Liu, Loks-Thompson, Marino, Cussons, Matthey, Mcloughlin, Mendolicchio, Merzic, Mitenkova, Moufarek, Oliveira, Oliveira, Openshaw, Pan, Pappu, Platonov, Purkiss, Reichert, Reid, Richemond, Roberts, Ruscoe, Elias, Sandars, Sawyer, Scholtes, Simmons, Slater, Soyer, Strathmann, Stys, Tam, Teplyashin, Terzi, Vercelli, Vujatovic, Wainwright, Wang, Wang, Wierstra, Williams, Wong, York, and Young]{SIMAteam}
SIMA Team, Maria~Abi Raad, Arun Ahuja, Catarina Barros, Frederic Besse, Andrew Bolt, Adrian Bolton, Bethanie Brownfield, Gavin Buttimore, Max Cant, Sarah Chakera, Stephanie C.~Y. Chan, Jeff Clune, Adrian Collister, Vikki Copeman, Alex Cullum, Ishita Dasgupta, Dario de~Cesare, Julia~Di Trapani, Yani Donchev, Emma Dunleavy, Martin Engelcke, Ryan Faulkner, Frankie Garcia, Charles Gbadamosi, Zhitao Gong, Lucy Gonzales, Kshitij Gupta, Karol Gregor, Arne~Olav Hallingstad, Tim Harley, Sam Haves, Felix Hill, Ed~Hirst, Drew~A. Hudson, Jony Hudson, Steph Hughes-Fitt, Danilo~J. Rezende, Mimi Jasarevic, Laura Kampis, Rosemary Ke, Thomas Keck, Junkyung Kim, Oscar Knagg, Kavya Kopparapu, Rory Lawton, Andrew Lampinen, Shane Legg, Alexander Lerchner, Marjorie Limont, Yulan Liu, Maria Loks-Thompson, Joseph Marino, Kathryn~Martin Cussons, Loic Matthey, Siobhan Mcloughlin, Piermaria Mendolicchio, Hamza Merzic, Anna Mitenkova, Alexandre Moufarek, Valeria Oliveira, Yanko Oliveira, Hannah Openshaw, Renke Pan, Aneesh Pappu, Alex
  Platonov, Ollie Purkiss, David Reichert, John Reid, Pierre~Harvey Richemond, Tyson Roberts, Giles Ruscoe, Jaume~Sanchez Elias, Tasha Sandars, Daniel~P. Sawyer, Tim Scholtes, Guy Simmons, Daniel Slater, Hubert Soyer, Heiko Strathmann, Peter Stys, Allison~C. Tam, Denis Teplyashin, Tayfun Terzi, Davide Vercelli, Bojan Vujatovic, Marcus Wainwright, Jane~X. Wang, Zhengdong Wang, Daan Wierstra, Duncan Williams, Nathaniel Wong, Sarah York, and Nick Young.
\newblock Scaling instructable agents across many simulated worlds.
\newblock \emph{arXiv}, 3 2024.
\newblock URL \url{http://arxiv.org/abs/2404.10179}.

\bibitem[Hawkins and Ahmad(2016)]{Hawkins2016WhyNeocortex}
Jeff Hawkins and Subutai Ahmad.
\newblock {Why Neurons Have Thousands of Synapses, a Theory of Sequence Memory in Neocortex}.
\newblock \emph{Frontiers in Neural Circuits}, 10, 2016.
\newblock ISSN 16625110.
\newblock \doi{10.3389/fncir.2016.00023}.

\bibitem[Hawkins et~al.(2017)Hawkins, Ahmad, and Cui]{Hawkins2017}
Jeff Hawkins, Subutai Ahmad, and Yuwei Cui.
\newblock {A Theory of How Columns in the Neocortex Enable Learning the Structure of the World}.
\newblock \emph{Frontiers in Neural Circuits}, 11\penalty0 (October):\penalty0 1--18, 2017.
\newblock ISSN 1662-5110.
\newblock \doi{10.3389/fncir.2017.00081}.
\newblock URL \url{http://journal.frontiersin.org/article/10.3389/fncir.2017.00081/full}.

\bibitem[Ahmad and Scheinkman(2019)]{Ahmad2019HowRepresentations}
Subutai Ahmad and Luiz Scheinkman.
\newblock {How Can We Be So Dense? The Robustness of Highly Sparse Representations}.
\newblock \emph{ICML 2019 Workshop on Uncertainty and Robustness in Deep Learning}, 2019.

\bibitem[Lewis et~al.(2019)Lewis, Purdy, Ahmad, and Hawkins]{Lewis2019LocationsCells}
Marcus Lewis, Scott Purdy, Subutai Ahmad, and Jeff Hawkins.
\newblock {Locations in the neocortex: A theory of sensorimotor object recognition using cortical grid cells}.
\newblock \emph{Frontiers in Neural Circuits}, 2019.
\newblock ISSN 16625110.
\newblock \doi{10.3389/fncir.2019.00022}.

\bibitem[Prasad et~al.(2020)Prasad, Carroll, and Sherman]{Prasad2020LayerTargets}
Judy~A. Prasad, Briana~J. Carroll, and S.~Murray Sherman.
\newblock {Layer 5 Corticofugal Projections from diverse cortical areas: Variations on a pattern of thalamic and extrathalamic targets}.
\newblock \emph{Journal of Neuroscience}, 40\penalty0 (30), 2020.
\newblock ISSN 15292401.
\newblock \doi{10.1523/JNEUROSCI.0529-20.2020}.

\bibitem[Savva et~al.(2019)Savva, Kadian, Maksymets, Zhao, Wijmans, Jain, Straub, Liu, Koltun, Malik, Parikh, and Batra]{habitat19iccv}
Manolis Savva, Abhishek Kadian, Oleksandr Maksymets, Yili Zhao, Erik Wijmans, Bhavana Jain, Julian Straub, Jia Liu, Vladlen Koltun, Jitendra Malik, Devi Parikh, and Dhruv Batra.
\newblock Habitat: {A} {P}latform for {E}mbodied {AI} {R}esearch.
\newblock In \emph{Proceedings of the IEEE/CVF International Conference on Computer Vision (ICCV)}, 2019.

\bibitem[Szot et~al.(2021)Szot, Clegg, Undersander, Wijmans, Zhao, Turner, Maestre, Mukadam, Chaplot, Maksymets, Gokaslan, Vondrus, Dharur, Meier, Galuba, Chang, Kira, Koltun, Malik, Savva, and Batra]{szot2021habitat}
Andrew Szot, Alex Clegg, Eric Undersander, Erik Wijmans, Yili Zhao, John Turner, Noah Maestre, Mustafa Mukadam, Devendra Chaplot, Oleksandr Maksymets, Aaron Gokaslan, Vladimir Vondrus, Sameer Dharur, Franziska Meier, Wojciech Galuba, Angel Chang, Zsolt Kira, Vladlen Koltun, Jitendra Malik, Manolis Savva, and Dhruv Batra.
\newblock Habitat 2.0: Training home assistants to rearrange their habitat.
\newblock In \emph{Advances in Neural Information Processing Systems (NeurIPS)}, 2021.

\bibitem[Puig et~al.(2023)Puig, Undersander, Szot, Cote, Partsey, Yang, Desai, Clegg, Hlavac, Min, Gervet, Vondrus, Berges, Turner, Maksymets, Kira, Kalakrishnan, Malik, Chaplot, Jain, Batra, Rai, and Mottaghi]{puig2023habitat3}
Xavi Puig, Eric Undersander, Andrew Szot, Mikael~Dallaire Cote, Ruslan Partsey, Jimmy Yang, Ruta Desai, Alexander~William Clegg, Michal Hlavac, Tiffany Min, Theo Gervet, Vladimir Vondrus, Vincent-Pierre Berges, John Turner, Oleksandr Maksymets, Zsolt Kira, Mrinal Kalakrishnan, Jitendra Malik, Devendra~Singh Chaplot, Unnat Jain, Dhruv Batra, Akshara Rai, and Roozbeh Mottaghi.
\newblock Habitat 3.0: A co-habitat for humans, avatars and robots, 2023.

\bibitem[Calli et~al.(2015)Calli, Singh, Walsman, Srinivasa, Abbeel, and Dollar]{YCB}
Berk Calli, Arjun Singh, Aaron Walsman, Siddhartha Srinivasa, Pieter Abbeel, and Aaron~M. Dollar.
\newblock The ycb object and model set: Towards common benchmarks for manipulation research.
\newblock In \emph{2015 International Conference on Advanced Robotics (ICAR)}, pages 510--517, 2015.
\newblock \doi{10.1109/ICAR.2015.7251504}.

\end{thebibliography}

\end{document}